\documentclass[11pt]{article}

\usepackage[final]{acl}
\usepackage{xcolor}
\usepackage{colortbl}

\usepackage{times}
\usepackage{latexsym}
\usepackage{amsmath}
\usepackage{algorithm}
\usepackage{algorithmic}
\usepackage{booktabs}
\usepackage{array}
\usepackage{tcolorbox}
\tcbuselibrary{skins, theorems, breakable, listings}
\usepackage{lipsum}
\usepackage{multicol}
\usepackage{verbatim}
\usepackage{mathptmx}
\usepackage{tabularx} 
\usepackage{listings}
\usepackage{graphicx}[2021/08/06]
\usepackage{longtable}
\usepackage{float} 
\usepackage{hyperref}
\hypersetup{
    colorlinks=true,
    urlcolor=blue,         
    linkcolor=darkblue,    
    citecolor=darkblue,    
    pdftitle={论文标题},
    pdfauthor={作者姓名},
    pdfsubject={论文主题},
    pdfkeywords={关键词1, 关键词2}
}
\newcolumntype{P}[1]{>{\raggedright\arraybackslash\vspace{0pt}}p{#1}}

\definecolor{mygreen}{RGB}{95, 180, 130}
\definecolor{mygreen2}{RGB}{186, 205, 190}
\definecolor{mygreen3}{RGB}{248, 250, 248}
\definecolor{tablehead}{RGB}{232, 238, 247}
\definecolor{toolblue}{RGB}{244, 248, 253}
\definecolor{nogray}{RGB}{247, 247, 247}
\definecolor{humanrow}{RGB}{242, 242, 242}
\newcolumntype{N}{>{\columncolor{nogray}}l}
\newcolumntype{U}{>{\columncolor{toolblue}}l}

\usepackage{multirow}
\usepackage{float}
\usepackage{subcaption}
\usepackage[T1]{fontenc}

\usepackage[utf8]{inputenc}

\usepackage{microtype}

\usepackage{inconsolata}
\usepackage{natbib}

\usepackage{graphicx}
\usepackage{acl}
\title{FinHopper: A Progressive Multi-Hop Benchmark for Complex Financial Agents with Step-wise Evaluation}

\author{
  \textbf{Yuxuan Wu\textsuperscript{1,2,*}},
  \textbf{Miaopei Lin\textsuperscript{2,*}},
  \textbf{Qiufang Ying\textsuperscript{2,*}},
  \textbf{Lin Dan\textsuperscript{2}},
  \textbf{Hao Zhao\textsuperscript{2}},
  \textbf{Qirui Liu\textsuperscript{1,2}},
\\
  \textbf{Yuteng Shen\textsuperscript{2}},
  \textbf{Shaohui Wu\textsuperscript{2}},
  \textbf{Weihong Luo\textsuperscript{2,\dag}},
  \textbf{Xiku Du\textsuperscript{2}},
  \textbf{Xing Tang\textsuperscript{3,\dag}},
  \textbf{Hao Chen\textsuperscript{2,\dag}}
\\
  \textsuperscript{1}South China University of Technology \\
  \textsuperscript{2}FiT, Tencent \\
  \textsuperscript{3}Shenzhen Technology University
}

\begin{document}
\maketitle

\let\thefootnote\relax
\footnotetext{$^{*}$\,Equal contribution.}
\footnotetext{$^{\dagger}$\,Corresponding Author.}

\begin{abstract}
Financial agents are expected to answer factual investment queries through open search, but financial multi-hop retrieval differs from general multi-hop QA in a crucial way: each intermediate answer is an exact typed constraint, such as an entity, date, event, or numerical indicator, that determines the next query.
This strict answer-conditioned dependency makes small intermediate errors cascade into plausible but wrong search trajectories.
We introduce \textbf{FinHopper}, a benchmark for complex financial agents with controlled reasoning depth from 1 to 6 hops, covering stocks, funds, and macroeconomic facts, with gold intermediate answers for process-level evaluation.
To construct FinHopper at scale, we design a state-transition-based synthesis framework that composes verified atomic financial QA patterns into dependency-preserving chains, followed by targeted in-synthesis verification, model-ensemble checking, and public-search validation by financial experts.
Experiments with 20 LLM-agent configurations show that public search tools are necessary but insufficient: performance drops sharply as hop depth increases, the best model remains far below human experts, and process-aware scoring reveals partial reasoning progress hidden by final-answer-only evaluation. 
The benchmark is available at \url{https://github.com/justajustin/FinHopper/}.

\end{abstract}

\section{Introduction}
Financial analysts rarely answer investment questions by retrieving a single fact. A typical workflow may first identify a target company from a comparison, use that company to query an event date, and then use the date to retrieve a market indicator or fund movement. LLM-based agents with search and tool-use abilities~\citep{nakano2021webgpt, yao2023react, schick2023toolformer} are increasingly expected to automate such evidence collection~\citep{wu2023bloomberggpt, nie2024survey, tongyidr}. This creates a financial multi-hop retrieval problem in which every hop must recover an exact, typed answer before the next search can even be posed correctly.

\begin{figure}[t]
    \centering
    \includegraphics[width=\columnwidth]{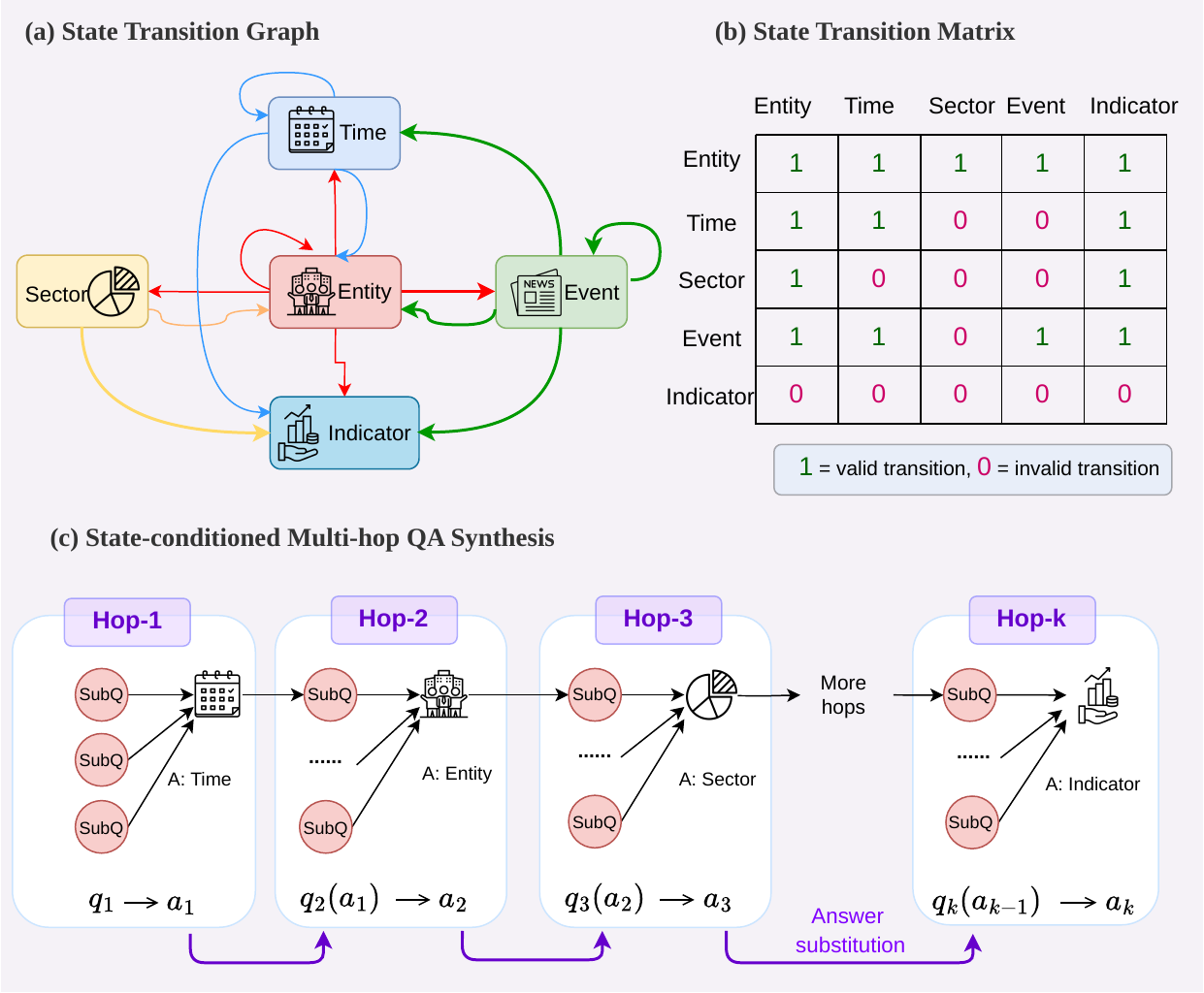}
    \caption{FinHopper abstracts financial multi-hop search as state-conditioned answer transitions, where each verified answer constrains the next hop through dependency-preserving answer substitution.}
    \label{fig:state_transition}
\end{figure}
\vspace{-1mm}

This structure is different from many general multi-hop QA or browsing tasks. In general domains, hops often gather complementary evidence, and approximate entity matching or broad web exploration can still lead to a usable answer. In finance, however, an intermediate answer becomes a hard query condition: a wrong company, reporting period, event date, or metric definition sends all later searches to the wrong state while the trajectory may still look coherent. Figure~\ref{fig:state_transition} abstracts this property as transitions among answer states, where only valid financial answer types can constrain subsequent hops. For example, identifying \textit{``the liquor company with the highest Q3 2024 revenue''} requires exact reported figures and period definitions, not a likely or famous brand.

This financial specificity motivates different benchmark design choices from general multi-hop data. A benchmark must control which answer states can be chained, preserve hidden intermediate answers, and verify that every hop remains factually grounded in public financial evidence. Existing financial benchmarks still leave this gap: FinQA~\citep{Finqa}, TAT-QA~\citep{zhu2021tat}, and ConvFinQA~\citep{convfinqa} emphasize numerical reasoning with provided evidence; FinAgentBench~\citep{financeagent} and related settings evaluate retrieval or tool use under more constrained formats; and FinSearchComp~\citep{finsearchcomp} moves closer to open financial search but is still primarily judged by final-answer correctness. Current benchmarks therefore rarely combine public-search reproducibility, strict answer-conditioned chaining, and process-level supervision.

We propose \textbf{FinHopper}, a benchmark designed around these financial multi-hop requirements. FinHopper contains questions from 1 to 6 reasoning hops across stocks, funds, and macroeconomic facts. Each item includes a final answer and step-wise gold intermediate answers, allowing evaluation to distinguish agents that fail immediately from those that retrieve correct intermediate evidence but break later in the chain. Its construction mirrors the motivation above: state-conditioned synthesis enforces valid financial dependencies, chain-merge verification prevents answer leakage or ambiguity, and expert public-search validation checks that released step answers are externally reproducible.

Our experiments show that FinHopper acts as a stress test for agentic financial retrieval. Without public search tools, all evaluated models nearly fail. With tools, performance improves substantially but still drops sharply as hop depth increases, revealing model-specific capability boundaries. Process-aware scoring further shows that some agents can make partial progress even when their final answers are wrong, which final-answer-only metrics would hide.

Our contributions are as follows:
\begin{enumerate}
\item \textbf{A benchmark for dependency-preserving financial search.} We introduce FinHopper, a multi-hop financial benchmark with controlled depth from 1 to 6 hops and hybrid coverage across stocks, funds, and macroeconomics.
\item \textbf{A scalable and verified construction framework.} We propose a state-transition-based synthesis method that composes atomic financial QA patterns into valid reasoning chains, with verification at both single-hop and chain-merge stages.
\item \textbf{A diagnostic evaluation protocol.} We provide step-wise gold answers and a process-aware scoring metric that localizes failures and exposes partial progress beyond binary final-answer accuracy.
\end{enumerate}

\section{Related Work}

\subsection{Multi-Hop QA and Web Agent Benchmarks}
Multi-hop QA and browsing benchmarks have driven progress in evidence retrieval, long-context reasoning, and tool-augmented interaction. HotpotQA~\citep{yang2018hotpotqa}, FEVER~\citep{thorne2018fever}, Wizard of Wikipedia~\citep{dinan2019wizard}, ELI5~\citep{fan2019eli5}, and KILT~\citep{petroni2021kilt} evaluate knowledge-intensive QA over retrieved or provided evidence. WebShop~\citep{yao2022webshop}, WebArena~\citep{zhou2023webarena}, Mind2Web~\citep{deng2023mind2web}, GAIA~\citep{mialon2023gaia}, and BrowseComp variants~\citep{browsecomp,browsecomp-ZH,browsecomp-Plus} further test web interaction and long-chain search. These benchmarks are not centered on financial answer-conditioned dependencies: many tasks reward successful navigation or approximate evidence gathering, whereas financial multi-hop retrieval requires exact entities, dates, metric definitions, and numerical values whose errors cascade across later hops.

\subsection{Financial QA and Search Benchmarks}
Financial benchmarks cover numerical reasoning, document understanding, and agentic decision making, but they do not fully capture open-domain sequential retrieval. FinQA~\citep{Finqa}, TAT-QA~\citep{zhu2021tat}, and ConvFinQA~\citep{convfinqa} focus on numerical reasoning over provided documents. FinAgent~\citep{zhang2024multimodal} and MarS~\citep{li2024mars} study LLM-based stock-market decision making with financial signals, while FinAgentBench~\citep{financeagent} introduces tool-augmented financial tasks under a more constrained retrieval setting. FinSearchComp~\citep{finsearchcomp} moves closer to open-domain financial search, but its complex setting is still primarily assessed with final-answer pass/fail outcomes. Some benchmarks provide rationales or step annotations, and Finance Agent Benchmark~\citep{bigeard2025finance} includes step-wise information, but they do not make verified intermediate answers central to benchmark construction and scoring. FinHopper instead targets public financial search where each intermediate answer becomes a typed condition for the next query, and its process-aware scoring diagnoses whether an agent fails in retrieval, state tracking, chain merging, or final synthesis.

\begin{figure*}[!t]
    \centering
    \includegraphics[width=0.9\textwidth]{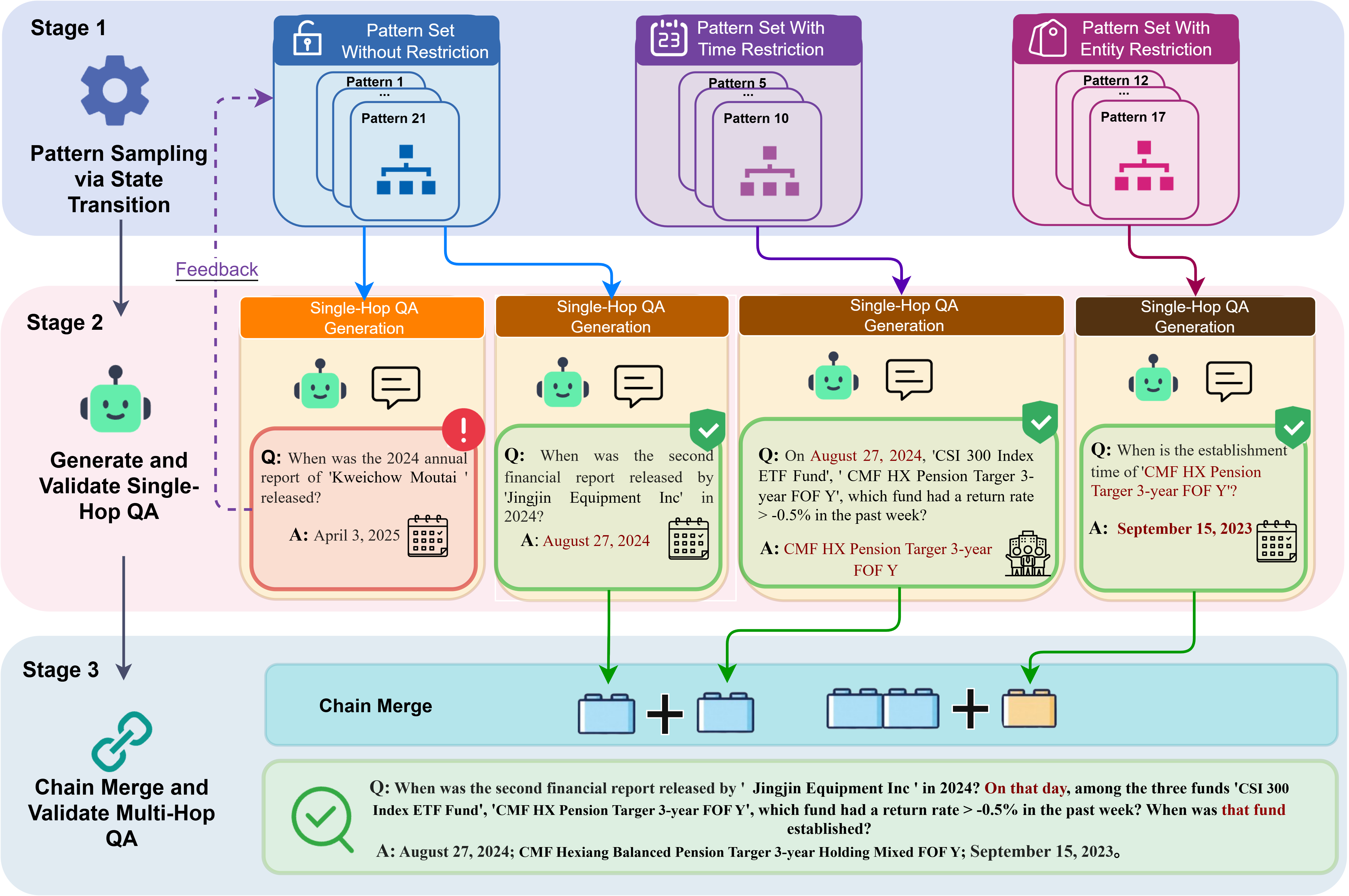}
    \caption{Fine-grained construction pipeline of FinHopper. The framework first samples reusable financial patterns according to valid answer-state transitions, then generates and validates single-hop QA pairs, and finally merges verified hops into multi-hop questions while preserving intermediate answers for process-level evaluation.}
    \label{fig:method_overview}
\end{figure*}

\section{Methodology}
\label{sec:methodology}
Building on Figure~\ref{fig:state_transition}, FinHopper composes verified atomic financial questions into dependency-preserving chains. Figure~\ref{fig:method_overview} summarizes the pipeline: state-conditioned pattern sampling, single-hop QA generation and validation, and chain merge with post-synthesis validation before release.

\subsection{Task Formulation}
\label{sec:task_formulation}
A FinHopper instance is a $K$-hop financial QA chain $\{(q_1,a_1), \ldots, (q_K,a_K)\}$, where $q_K$ is the released user-facing question and $a_K$ is the final answer. The intermediate answers $\{a_1,\ldots,a_{K-1}\}$ are not exposed to the model during evaluation, but they are retained as gold process annotations. The defining property is strict dependency: answer $a_t$ must provide a concrete constraint, such as an entity, date, sector, event, or indicator, that is required to formulate or solve the next hop $q_{t+1}$.

We distinguish three levels of granularity. A \textbf{simple question} queries one elementary financial relation, such as an entity-time-indicator fact or an event date. A \textbf{single-hop question} may contain multiple simple sub-queries that can be solved in parallel, such as comparing several entities under the same time and indicator. A \textbf{multi-hop question} links single-hop questions sequentially through answer substitution, so that each new hop depends on the verified answer of the previous hop.

\subsection{Atomic Financial Patterns}
\label{sec:atomic_patterns}
To make synthesis controllable, we define a pattern library for generating single-hop QA pairs. Each pattern specifies the input components, the required computation, and the answer type. Across financial domains, these components can be represented by five answer states:
\[
\text{S} = \{\text{Entity, Time, Sector, Event, Indicator}\}.
\]
We design 21 patterns covering entity-time comparison, event-based retrieval, and entity-relationship queries. Pattern variants combine the question component type, computation type (e.g., ranking, filtering, comparison, or extraction), and answer component type, which determines how the next hop can be sampled. The full taxonomy is provided in Appendix~\ref{appendix:patterns}.

\subsection{State-Conditioned Chain Synthesis}
\label{sec:state_transition}
Not every answer type can validly constrain every subsequent question. For example, an entity answer can lead to questions about its sector, event timeline, or indicators, while an indicator value usually functions as a terminal answer. We therefore model chain construction as a random walk over an answer-state transition graph.

We define a state transition graph $G = (\text{S}, \text{E})$, where each directed edge $(s_i \to s_j) \in \text{E}$ indicates that an answer of state $s_i$ can condition a following question whose answer state is $s_j$, as shown in Figure~\ref{fig:state_transition}. The transition probability from state $s_i$ to $s_j$ is defined by the normalized transition matrix $\mathbf{P} \in \{0,1\}^{|\text{S}| \times |\text{S}|}$:

\begin{equation}
P_{ij} = \begin{cases} 
\frac{1}{|D(s_i)|} & \text{if } (s_i \to s_j) \in \text{E} \\ 
0 & \text{otherwise} 
\end{cases}
\end{equation}

where $D(s_i) = \{s_j : (s_i \to s_j) \in \text{E}\}$ is the set of valid outgoing states. Starting from an initial state $s_0 \sim \pi_0$, we sample a target depth $K$ and generate a valid trajectory $(s_0,\ldots,s_K)$ using $\mathbf{P}$. Each sampled state selects an eligible pattern and data pool; after verification, its answer is substituted into the next hop as an explicit constraint. Detailed transition examples are listed in Table~\ref{tab:state_transition_examples}.

\subsection{Chain Generation and Merge}
\label{sec:generation}

\subsubsection{Single-Hop Generation}
\label{sec:single_hop}
Given a sampled pattern, an LLM generator equipped with internal financial database tools retrieves candidate factual evidence and instantiates a single-hop QA pair. Each pair must pass single-hop verification before it can be merged into a longer chain.

\subsubsection{Chain Merge}
\label{sec:chain_merge}

After a single-hop QA passes verification, it is merged with the existing chain through \textbf{answer substitution}: the previous hop's answer is embedded into the next hop as a required constraint. The released question hides intermediate answers, while the benchmark stores them as process-level gold references. Chain-merge verification ensures that the new question is self-contained, unambiguous, and faithful to the underlying evidence.

\subsection{Quality Control}
\label{sec:quality_control}

FinHopper uses quality control at two stages: in-synthesis verification filters errors during generation, and post-synthesis verification validates released items independently from the construction process.

\subsubsection{In-Synthesis Verification}
In-synthesis verification checks both atomic QA pairs and chain-merge operations before they enter the final benchmark, preventing factual errors from accumulating across hops. The detailed verification flow is shown in Appendix~\ref{appendix:in_synthesis_verification_details}.

\paragraph{Single-Hop Verification}
Single-hop verification re-derives tool parameters from the question text alone, retrieves evidence, and checks answer consistency. Failed or ambiguous samples are regenerated rather than merged.

\paragraph{Chain-Merge Verification}
For chain merge, separate critics check the question and answer: the question critic prevents leakage, ambiguity, and logical drift, while the answer critic checks completeness and faithfulness to evidence. We sample parallel merge candidates and retain the best validated one. Detailed criteria and prompts are provided in Appendices~\ref{appendix:Guide For Chain-Merge Verification} and~\ref{appendix: Prompts of Quality Control}.

\subsubsection{Post-Synthesis Verification}
After synthesis, we apply two independent checks before releasing the benchmark: model-ensemble answer verification and human verification through public search.

\paragraph{Answer Correctness Verification}
We employ an ensemble of five frontier LLMs under a ReAct framework~\citep{yao2023react}. At this construction-verification stage, models query the internal financial database to derive reasoning chains and final answers. A sample is accepted only when a strict majority ($\geq 3$ agents) agree; disagreements are sent to expert arbitration. Implementation details are provided in Appendix~\ref{appendix:Prompt of React Framework}.

\paragraph{Human Verification}
As a final safeguard, 16 financial experts verified every released step-level answer through public search engines rather than the internal database. We also sampled 30\% of synthesized items from each complexity level for double-blind review and senior-expert adjudication. The 95\% pass rate confirms alignment between automated and expert public-search checks.

\section{Experiments}

We conduct an evaluation on FinHopper to assess the capability of LLM agents in addressing financial tasks. Our evaluation focuses on four key aspects: the necessity of tools, the human-AI performance gap, the impact of reasoning depth, and inference efficiency.

\begin{figure*}[t]
    \centering
    \begin{subfigure}[b]{0.32\textwidth}
        \centering
        \includegraphics[width=\textwidth]{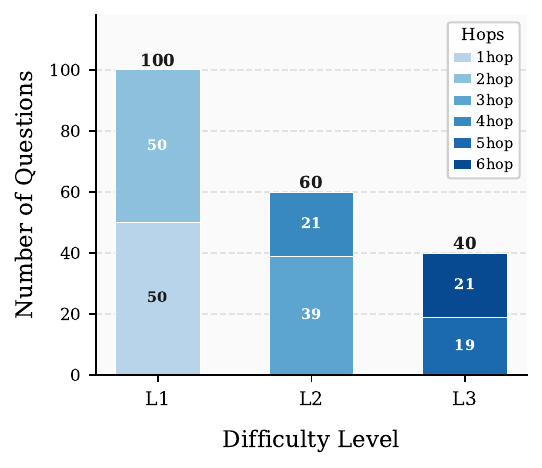}
        \caption{Difficulty \& Hops}
        \label{fig:benchmark_hop_distribution}
    \end{subfigure}
    \hfill
    \begin{subfigure}[b]{0.32\textwidth}
        \centering
        \includegraphics[width=\textwidth]{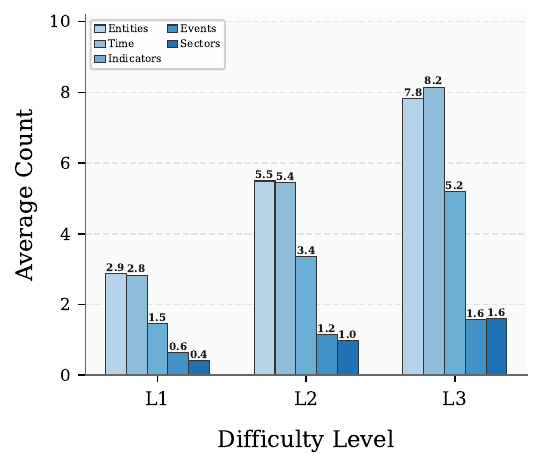} 
        \caption{Question Complexity}
        \label{fig:etm_by_level}
    \end{subfigure}
    \hfill
    \begin{subfigure}[b]{0.32\textwidth}
        \centering
        \includegraphics[width=\textwidth]{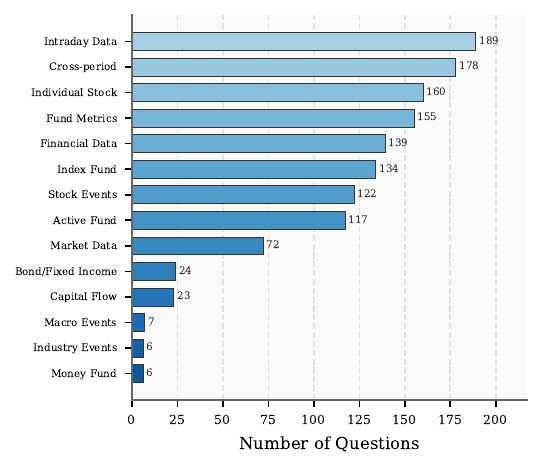}
        \caption{Domain Distribution}
        \label{fig:domain_bar}
    \end{subfigure}
    \caption{FinHopper provides a controlled difficulty gradient: hop depth increases from L1 to L3, question components grow denser with difficulty, and items cover diverse financial domains across stocks, funds, and macroeconomics.}
    \label{fig:benchmark_statistics}
\end{figure*}

\subsection{Benchmark Statistics}

Figure~\ref{fig:benchmark_statistics} summarizes FinHopper. The benchmark contains 200 questions with a controlled difficulty gradient: L1 covers 1--2 hops, L2 covers 3--4 hops, and L3 covers 5--6 hops. As difficulty increases, question components also become denser: L3 questions involve 2.7$\times$ more entities and 2.9$\times$ more time references than L1. The benchmark spans 14 financial domains, including intraday data, cross-period analysis, individual stocks, fund metrics, and macroeconomics. FinHopper includes Chinese and English versions with the same reasoning chains and step-level gold answers; the main large-scale evaluation uses the Chinese version, with representative English results in Section~\ref{sec:english_rep_eval}.

\begin{table}[t]
\centering
\small
\caption{Hop-wise composition of FinHopper. Each $K$-hop item contains $K$ step-level gold answers, yielding 572 step-level checking points in total. These step-level annotations are diagnostic signals rather than independent test samples.}
\label{tab:hop_distribution}
\setlength{\tabcolsep}{4pt}
\begin{tabular}{lcccccc}
\toprule
\textbf{Hop} & 1 & 2 & 3 & 4 & 5 & 6 \\
\midrule
\textbf{Level} & L1 & L1 & L2 & L2 & L3 & L3 \\
\textbf{\#Items} & 50 & 50 & 39 & 21 & 19 & 21 \\
\textbf{\#Step answers} & 50 & 100 & 117 & 84 & 95 & 126 \\
\bottomrule
\end{tabular}
\end{table}

\subsection{Evaluation of FinHopper}
Financial tasks require multi-hop reasoning, so binary final-answer metrics cannot capture partial progress or hallucinated trajectories. We therefore use a two-stage hierarchical score. First, we set $s_{final}=1$ if the generated final answer matches the gold answer after normalizing units, dates, and formatting; otherwise $s_{final}=0$. If the final answer is wrong, we evaluate whether the response recovers the $N$ gold intermediate answers $\text{K}=\{k_{1},k_{2},...,k_{N}\}$:
{\setlength{\abovedisplayskip}{4pt}
\setlength{\belowdisplayskip}{4pt}
\[
s_{process}=\alpha \cdot \frac{1}{N} \sum_{i=1}^{N} I(k_{i} \preceq \text{A})
\]}
where $k_i \preceq \text{A}$ means that the response entails the gold intermediate step $k_i$, and $\alpha=0.5$ caps partial credit for incorrect final answers. The final score is:
{\setlength{\abovedisplayskip}{4pt}
\setlength{\belowdisplayskip}{4pt}
\[
S(\text{A})=s_{final}+(1-s_{final}) \cdot s_{process}
\]}
Correct final answers receive 1.0, while incorrect final answers receive 0.0--0.5 based on process validity. Evaluation prompts are provided in Appendix~\ref{appendix:Prompt of Evaluation}.

\begin{table*}[h!]
    \centering 
    \caption{Performance of various models and human on FinHopper. All scores are reported in \%.} 
    \label{tab:Performance of various models and human} 
    \small
    \setlength{\tabcolsep}{4pt}
\begin{tabular}{l|NN|UUUUUUUUU}
\hline
\rowcolor{tablehead}
\multicolumn{1}{c|}{} & \multicolumn{2}{c|}{\textbf{without tools}}                                  & \multicolumn{9}{c}{\textbf{use tools}}                                                                                                                                                                                                      \\ \cline{2-12} 
\rowcolor{tablehead}
\multicolumn{1}{c|}{\textbf{model}}                  & \textbf{\begin{tabular}[c]{@{}l@{}}Mic \\ Avg.\end{tabular}} & \textbf{Time} & \textbf{1hop} & \textbf{2hop} & \textbf{3hop} & \textbf{4hop} & \textbf{5hop} & \textbf{6hop} & \textbf{\begin{tabular}[c]{@{}l@{}}Mic \\ Avg.\end{tabular}} & \textbf{\begin{tabular}[c]{@{}l@{}}Mac \\ Avg.\end{tabular}} & \textbf{Time} \\ \hline
\rowcolor{humanrow}
Human Performance                                    & 0                                                            & 5             & 94.0          & 90.0          & 84.6          & 73.4          & 65.9          & 64.4          & 81.7                                                         & 78.7                                                         & 324           \\ \hline
Claude-4.5-Sonnet-thinking                           & 2.5                                                          & 18            & 81.3          & \textbf{72.5} & \textbf{70.2} & 63.5          & \textbf{54.1} & 31.1          & \textbf{67.3}                                                & \textbf{62.1}                                                & 31            \\
Claude-4.5-Sonnet-no-thinking                        & 1.0                                                          & 10            & 79.6          & 70.7          & 63.2          & 60.0          & 49.6          & 25.6          & 63.7                                                         & 58.1                                                         & \textbf{24}   \\
Gemini-3-Pro-thinking                                & 3.2                                                          & 18            & 80.0          & 69.1          & 69.5          & 60.1          & 52.2          & 26.5          & 65.0                                                         & 59.6                                                         & 34            \\
Gemini-3-Pro-no-thinking                             & 2.8                                                          & 14            & 81.2          & 70.3          & 65.5          & 53.0          & 44.2          & 20.2          & 62.7                                                         & 55.7                                                         & 25            \\
GPT-5.2-thinking                                     & 4.2                                                          & 30            & 83.4          & 69.6          & 56.0          & 53.4          & 45.8          & 19.2          & 61.3                                                         & 54.6                                                         & 121           \\
GPT-5.2                                              & 1.4                                                          & 27            & 86.8          & 64.4          & 50.2          & 49.2          & 44.7          & 12.5          & 58.5                                                         & 51.3                                                         & 108           \\
Doubao-seed-1-8-251215-thinking                      & 1.8                                                          & 15            & 79.0          & 61.3          & 60.2          & \textbf{63.9} & 46.6          & 28.8          & 61.1                                                         & 56.6                                                         & 34            \\
Doubao-seed-1-8-251215                               & 3.2                                                          & 14            & 78.1          & 63.4          & 58.4          & 57.0          & 41.2          & 23.2          & 59.2                                                         & 53.6                                                         & 30            \\
DeepSeek-V3.2-thinking                               & 2.0                                                          & 24            & 86.3          & 72.3          & 65.7          & 34.8          & 10.7          & 9.3           & 58.1                                                         & 46.5                                                         & 42            \\
DeepSeek-V3.2-no-thinking                            & 4.2                                                          & 19            & \textbf{88.5} & 69.2          & 63.1          & 34.5          & 9.2           & 8.0           & 57.1                                                         & 45.4                                                         & 34            \\
Kimi-k2-thinking                                     & 3.4                                                          & 23            & 67.2          & 58.5          & 61.1          & 53.2          & 48.2          & \textbf{38.0} & 57.5                                                         & 54.4                                                         & 203           \\
Kimi-k2                                              & 7.0                                                          & 20            & 68.0          & 50.9          & 59.7          & 49.6          & 44.0          & 36.1          & 54.6                                                         & 51.4                                                         & 187           \\
HunYuan-2.0-thinking                                 & 4.0                                                          & 26            & 80.1          & 60.9          & 55.5          & 51.5          & 31.5          & 26.4          & 57.3                                                         & 51.0                                                         & 46            \\
HunYuan-2.0-instruct                                 & 3.2                                                          & 19            & 76.5          & 58.0          & 49.2          & 47.5          & 24.8          & 20.0          & 52.7                                                         & 46.0                                                         & 41            \\
Grok-4.1-thinking                                    & 3.0                                                          & 16            & 66.2          & 57.7          & 53.0          & 50.3          & 35.3          & 25.8          & 52.7                                                         & 48.1                                                         & 57            \\
Grok 4.1                                             & 6.2                                                          & 14            & 63.0          & 51.9          & 49.5          & 46.1          & 27.0          & 18.7          & 42.7                                                         & 42.7                                                         & 42            \\
Qwen-max-20250923-thinking                           & 3.2                                                          & 14            & 63.0          & 54.7          & 53.8          & 55.9          & 34.5          & 32.4          & 52.5                                                         & 49.1                                                         & 47            \\
Qwen-max-20250923-no-thinking                        & 1.0                                                          & \textbf{6}    & 61.3          & 44.5          & 53.4          & 50.8          & 30.6          & 30.5          & 48.3                                                         & 45.2                                                         & 30            \\
GLM-4.7-thinking                                     & 8.1                                                          & 71            & 76.1          & 47.4          & 50.7          & 43.6          & 22.7          & 14.2          & 49.0                                                         & 42.5                                                         & 92            \\
GLM-4.7-no-thinking                                  & 6.4                                                          & 57            & 75.2          & 45.5          & 45.3          & 38.5          & 18.7          & 11.3          & 48.8                                                         & 40.9                                                         & 73            \\ \hline
\end{tabular}
\end{table*}

\subsection{Experimental Setup}
We evaluate 10 representative frontier model families on the Chinese version of FinHopper: Gemini 3 Pro~\citep{google2025gemini3}, Claude 4.5 Sonnet~\citep{anthropic2025claudesonnet45}, Doubao-seed-1-8-251215~\citep{bytedance2025seed18}, GPT-5.2~\citep{openai2025gpt52}, DeepSeek-V3.2~\citep{deepseek2025v32}, HunYuan-2.0~\citep{tencent2026hunyuan20}, Kimi K2 and K2 Thinking~\citep{kimiteam2025k2,kimi2025k2thinking}, GLM-4.7~\citep{zai2025glm47}, qwen3-max-2025-09-23~\citep{alibaba2025qwen3max}, and Grok 4.1~\citep{xai2025grok41}. Models answer with public search engines under the ReAct framework~\citep{yao2023react}, with a maximum iteration budget of 42; the internal financial database is used only for data construction and is never exposed during evaluation. We test both "thinking" and "no-thinking" modes, and use 20 independent financial experts under the same public-search setting as the human upper bound. ReAct prompts are provided in Appendix~\ref{appendix:Prompt of React Framework}.

Because simple questions (1--2 hops) constitute 50\% of the data, we report two metrics to avoid biasing results.

\textbf{Micro-Average Score}: This measures overall performance weighted by the data distribution. Let \( Score_k \) denote the score at hop \( k \), and \( w_k \) be the proportion of \( k \)-hop questions:
{\setlength{\abovedisplayskip}{4pt}
\setlength{\belowdisplayskip}{4pt}
\[
Score_{\text{micro-avg}} = \sum_{k=1}^{6} w_k \cdot Score_k
\]}

\textbf{Macro-Average Score}: This treats all hop depths equally and better reflects robustness on long-chain tasks:
{\setlength{\abovedisplayskip}{4pt}
\setlength{\belowdisplayskip}{4pt}
\[
Score_{\text{macro-avg}} = \frac{1}{6} \sum_{k=1}^{6} Score_k
\]}
We also measure average inference latency in seconds per query.

\begin{table*}[t]
    \centering
    \caption{Performance of representative models on FinHopper (English). All scores are reported in \%.}
    \label{tab:english_rep_eval}
    \small
    \setlength{\tabcolsep}{4pt}
\begin{tabular}{l|NN|UUUUUUUUU}
\hline
\rowcolor{tablehead}
\multicolumn{1}{c|}{} & \multicolumn{2}{c|}{\textbf{without tools}}                                  & \multicolumn{9}{c}{\textbf{use tools}}                                                                                                                                                                                                      \\ \cline{2-12} 
\rowcolor{tablehead}
\multicolumn{1}{c|}{\textbf{model}}                  & \textbf{\begin{tabular}[c]{@{}l@{}}Mic \\ Avg.\end{tabular}} & \textbf{Time} & \textbf{1hop} & \textbf{2hop} & \textbf{3hop} & \textbf{4hop} & \textbf{5hop} & \textbf{6hop} & \textbf{\begin{tabular}[c]{@{}l@{}}Mic \\ Avg.\end{tabular}} & \textbf{\begin{tabular}[c]{@{}l@{}}Mac \\ Avg.\end{tabular}} & \textbf{Time} \\ \hline
Gemini-3-Pro-thinking                                & 3.5                                                          & \textbf{17}   & 81.5          & 68.3          & \textbf{70.8} & \textbf{60.8} & \textbf{51.5} & 26.0          & \textbf{65.2}                                                & \textbf{59.8}                                                & \textbf{33}   \\
GPT-5.2-thinking                                     & 4.8                                                          & 29            & \textbf{84.8} & \textbf{69.0} & 57.2          & 54.0          & 45.2          & 19.8          & 61.7                                                         & 55.0                                                         & 118           \\
GLM-4.7-thinking                                     & 7.5                                                          & 74            & 75.0          & 48.2          & 51.0          & 42.8          & 22.2          & 14.8          & 48.8                                                         & 42.3                                                         & 95            \\
Kimi-k2-thinking                                     & 3.0                                                          & 25            & 66.0          & 59.2          & 60.5          & 52.5          & 48.8          & \textbf{38.2} & 57.3                                                         & 54.2                                                         & 208           \\ \hline
\end{tabular}
\end{table*}

\begin{figure*}[t]
    \centering
    \includegraphics[width=1.0\textwidth]{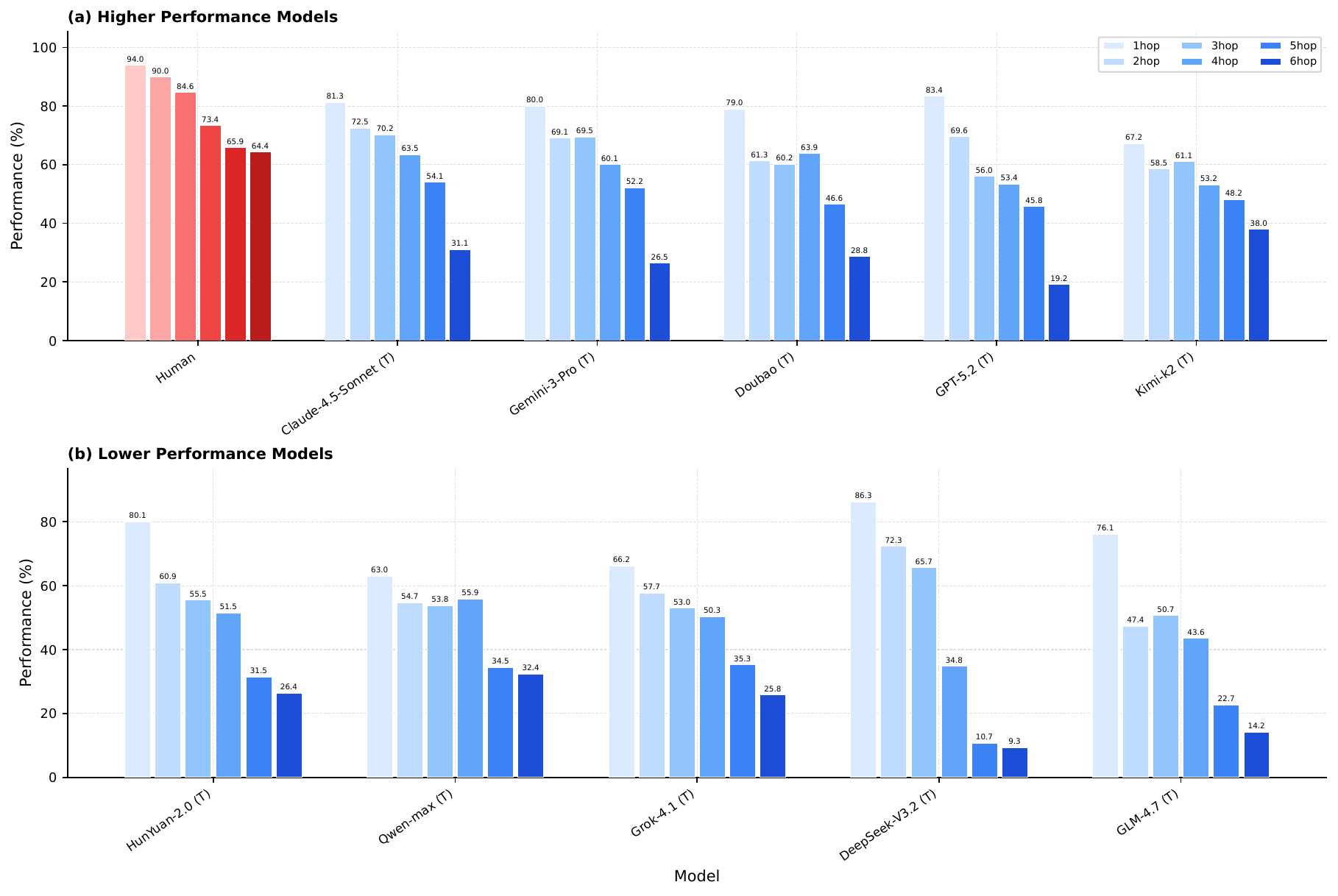} 
    \caption{Model performance across 1--6 hop tasks, grouped by thinking mode. The curves expose model-specific capability boundaries: most agents remain competitive on shallow tasks but degrade sharply once the reasoning chain exceeds their effective search and state-tracking depth.} 
    \label{fig:performance trend}
\end{figure*}

\begin{figure*}[t]
    \centering 
    \includegraphics[width=0.8\textwidth]{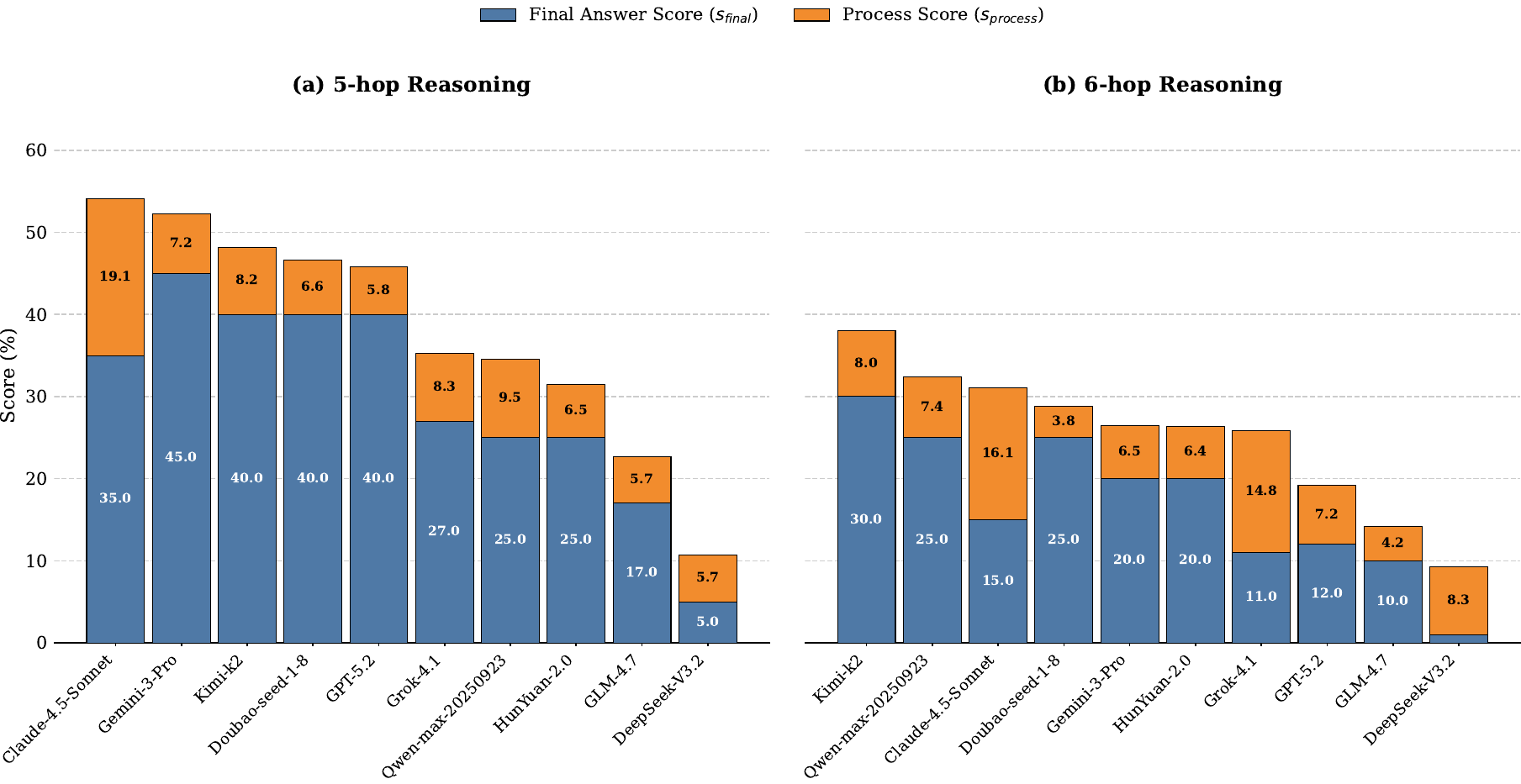} 
    \caption{Breakdown of model performance into Final Answer Score($s_{final}$) and Process Score ($s_{process}$). The left and right panels show results for 5-hop and 6-hop questions, respectively. } 
    \label{fig:final_vs_process}
\end{figure*}

\begin{figure}[h!]
    \centering 
    \includegraphics[width=0.48\textwidth]{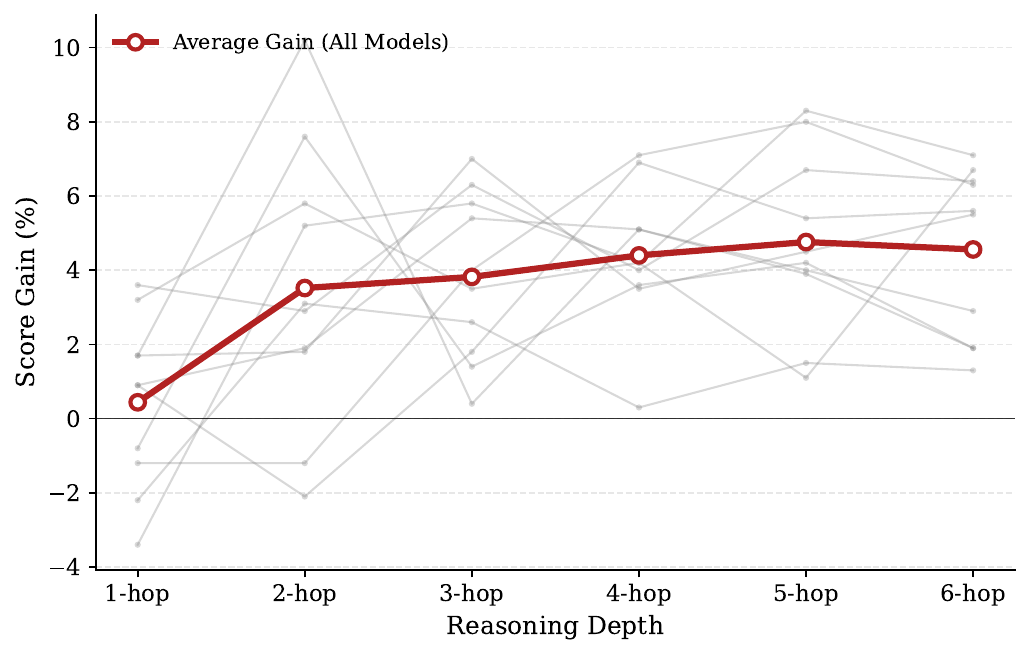} 
    \caption{Score gain of Thinking Mode across tasks with depth from 1 to 6 hops. The grey lines represent the score trajectories of individual models, while the bold red line denotes the average gain of all models.} 
    \label{fig:thinking_gain_trend_analysis}
\end{figure}

\subsection{Main Result}
We report the overall performance on the Chinese version in Table~\ref{tab:Performance of various models and human}, with English-version representative results provided in Table~\ref{tab:english_rep_eval}. A sharp contrast exists between the "without tools" and "use tools" settings. Without access to public search tools, all models fail almost completely, scoring below 8.1\%. 
With search-tool augmentation, 
Claude-4.5-Sonnet-thinking achieves the top Micro Average score of 67.3\%,
but there remains \textbf{a significant gap to human experts} (81.7\%).
This gap expands as the question difficulty increases. For instance, on complex 6-hop tasks, humans maintain 64.4\% accuracy, while the best model achieves only 31.1\%. A detailed case study is provided in Appendix~\ref{appendix:case study}.
To assess the stability of the reported results under repeated evaluations, we report mean and standard deviation for both Micro- and Macro-Average scores in Appendix~\ref{appendix:stability_analysis}. The standard deviations remain within 0.7--1.8 points, indicating that the main ranking trends are stable despite the moderate benchmark size.

Furthermore, we present three key findings:

\textbf{(1) Each model has a specific "agentic capability boundary"}. We observe a sharp cliff-like performance decline for most models once the question depth reaches a certain hop. As shown in Figure~\ref{fig:performance trend}, DeepSeek-V3.2 performs well on simple tasks but the score drops by half after 4-hop. Similarly, Gemini-3-Pro maintains high accuracy but eventually hits its boundary at 6-hop.

\textbf{(2) Models with "thinking" capabilities demonstrate superior robustness compared to standard counterparts.} Figure~\ref{fig:thinking_gain_trend_analysis} illustrates this performance gain. The benefit of the thinking mode becomes more evident as reasoning difficulty increases.
This suggests that the main benefit of reasoning-enhanced inference is not memorizing more financial facts, but maintaining a more reliable execution trace. In long-chain financial search, a model must decompose the user query into ordered sub-goals, preserve typed constraints such as entities, dates, sectors, and indicators, and revise its search plan when an intermediate result conflicts with later evidence. Thinking models appear to be more robust at these planning and state-tracking operations, which explains why the average gain increases from shallow to deeper hops. However, the gain is not uniform across all depths or models: on simple 1--2 hop questions, the additional reasoning budget sometimes provides limited benefit because the main bottleneck is factual retrieval rather than long-horizon control. While the "thinking" mode enhances accuracy, it incurs a computational cost. As shown in the Time column in Table~\ref{tab:Performance of various models and human}, reasoning-enhanced models generally require longer inference times (e.g., GPT-5.2-thinking at 121s vs. GPT-5.2 at 108s).

\textbf{(3) Step-by-step grading offers more precise assessment of agent capabilities than the binary final answer score }, especially for complex tasks (5-6 hops), as shown in Figure~\ref{fig:final_vs_process}. A binary score is not able to distinguish answers with different partial success. For instance, on 6-hop tasks, DeepSeek-V3.2 scores only 1.0 on $s_{final}$ but achieves 8.3 on $s_{process}$, indicating the model can solve intermediate steps even if it fails the final check. Thus, step-by-step grading is vital for capturing the true potential of the agents.

\subsection{Representative English-Version Evaluation}
\label{sec:english_rep_eval}
FinHopper also includes an English version translated from the Chinese benchmark with the same reasoning chains, final answers, and step-level gold references. Because full tool-augmented evaluation incurs substantial API and search costs, we evaluate the English version on four representative thinking models. As shown in Table~\ref{tab:english_rep_eval}, the English-version results follow the same broad trend as the Chinese main evaluation, with strong performance on lower-hop tasks and a clear decline as reasoning depth increases.

\subsection{Failure Analysis}
\label{sec:failure_analysis}

To better understand the capability boundary revealed by FinHopper, we sample successful and failed model trajectories across difficulty levels and annotate the first major error that causes the reasoning chain to deviate. The failures mainly fall into four categories: decomposition and planning failures (38\%), tool misuse (31\%), hallucination propagation (17\%), and information extraction failures (14\%). These patterns suggest that performance drops beyond 4 hops not because models lack isolated financial facts, but because longer chains amplify small errors in planning, tool grounding, evidence extraction, and state tracking into final-answer failures. Detailed examples and qualitative analysis are provided in Appendix~\ref{appendix:case study}.

\section{Conclusion}

In this paper, we address the lack of benchmarks for agents' capacity in solving complex financial tasks requiring strict sequential dependency and disambiguation of professional metrics by introducing \textbf{FinHopper}. It is constructed through a state-transition-based synthesis framework with verified intermediate answers. Our evaluation reveals that while the top-performing model, Claude-4.5-Sonnet-thinking, demonstrates improved robustness, a substantial performance gap remains compared to human experts, particularly on tasks exceeding 4 hops. Furthermore, we provide a step-by-step diagnosis tool with validated intermediate gold reference answers. In future work, we plan to expand the benchmark boundaries by incorporating broader heterogeneous data sources and synthesizing reasoning chains of higher complexity, continually challenging the limits of agentic reasoning in professional financial domains.

\section{Limitations}
Despite FinHopper's advancements, limitations remain. \textbf{Language and Market Coverage.} Although FinHopper provides both Chinese and English versions, the main large-scale evaluation is conducted on Chinese questions, and the data predominantly stems from the Chinese market, leaving broader generalization to global markets to be further verified. \textbf{Naturalness of Queries.} Although auto-generated queries undergo human verification, they still lack the ambiguity and informality typical of real-world questions. \textbf{Scope and Modality.} The benchmark focuses on factual reasoning, currently excluding subjective tasks like sentiment analysis or multi-modal interpretation.

\section{Ethical Review}
\textbf{Data Collection and Privacy.} 
FinHopper is constructed solely from publicly available sources (regulatory filings, market data, official disclosures) and contains no personally identifiable information (PII). The released benchmark contains the query, final gold answer, and step-wise intermediate answers required for process evaluation. All collection processes comply with respective terms of service and data protection regulations.

\textbf{Intended Use and Potential Misuse.} 
Designed for evaluating AI reasoning, FinHopper mitigates misuse risks (e.g., market manipulation) by focusing on historical data and reasoning patterns rather than real-time trading signals. We advocate for responsible research in compliance with financial regulations.

\textbf{Bias Considerations.} 
To address potential biases in financial data (e.g., over-representation of large-cap firms), we employed stratified sampling across company sizes, sectors, and time periods. Users should consider these inherent limitations when interpreting results.

\textbf{Human Evaluation.} 
Expert evaluations were conducted with informed consent from voluntary participants. No sensitive personal data was collected during the review process.



\bibliography{custom}

\appendix

\section{In-Synthesis Verification Details}
\label{appendix:in_synthesis_verification_details}
\begin{figure}[t]
    \centering
    \includegraphics[width=\columnwidth]{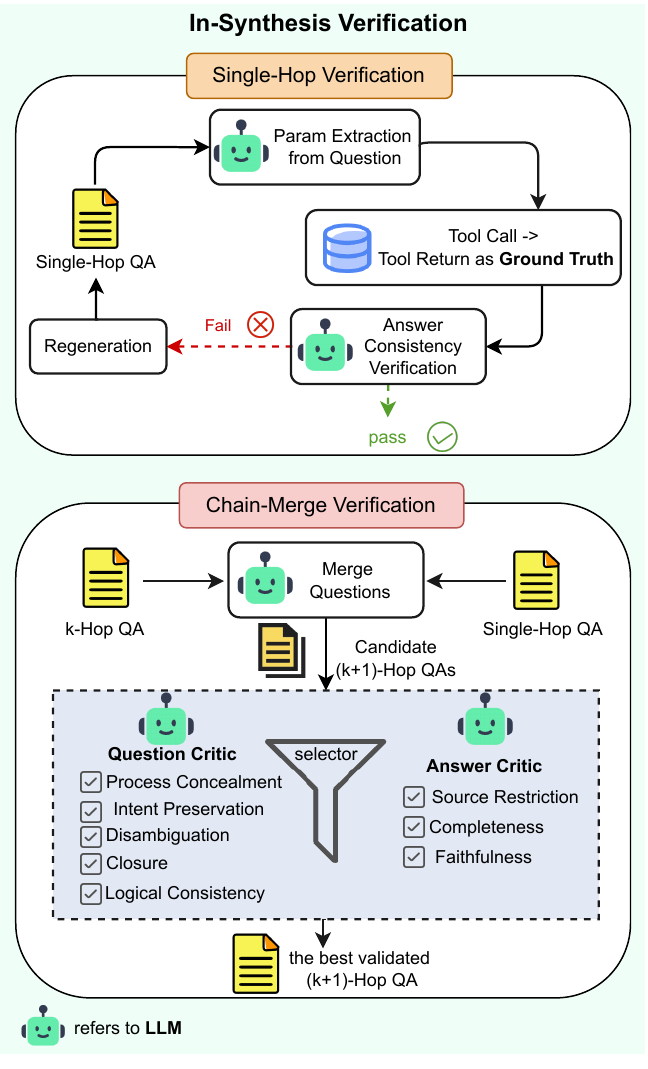} 
    \caption{In-synthesis verification independently re-derives single-hop answers and critiques chain-merged questions before they are accepted into longer FinHopper instances.} 
    \label{fig:in_synthesis_verification}
\end{figure}

Figure~\ref{fig:in_synthesis_verification} illustrates the two filters used during benchmark construction. Single-hop verification independently re-derives answers from question text, while chain-merge verification critiques candidate merged questions and answers before accepting a longer QA chain.

For single-hop verification, the verifier receives only the generated question, re-extracts tool parameters, invokes the corresponding financial tools, and checks whether the generated answer is semantically consistent with retrieved evidence. Samples whose required parameters cannot be reliably extracted are treated as ambiguous and regenerated. For chain-merge verification, the question critic checks process concealment, intent preservation, disambiguation, closure, and logical consistency, while the answer critic checks evidence restriction, completeness, and faithfulness. These checks target the key risks of financial multi-hop synthesis: leaking intermediate answers, drifting away from the original constraint, or propagating unsupported evidence across hops.

\section{Data Pool Definitions}
\label{appendix:data_pools}
As illustrated in Table~\ref{tab:data_pools_appendix}, the input states are categorized into four distinct dimensions: \textit{Entity}, \textit{Time}, \textit{Sector}, and \textit{Event}.

\section{Experiment Details on FinHopper}
\subsection{Model Performance with Increasing Task Difficulty}
\label{appendix:Model Performance with Increasing Task Difficulty}
The detailed 1--6 hop performance trends are discussed in the main text alongside Figure~\ref{fig:performance trend}.

\subsection{Stability Analysis}
\label{appendix:stability_analysis}
Table~\ref{tab:stability_analysis} reports the mean and standard deviation of Micro- and Macro-Average scores under repeated evaluations. The overall fluctuations are small, supporting the stability of the main conclusions.

\begin{table}[htbp]
\centering
\small
\caption{Stability analysis on FinHopper. Scores are reported as mean $\pm$ standard deviation in \%.}
\label{tab:stability_analysis}
\setlength{\tabcolsep}{5pt}
\begin{tabular}{lcc}
\toprule
\textbf{Model} & \textbf{Mic Avg.} & \textbf{Mac Avg.} \\
\midrule
Human Performance & $81.7 \pm 1.6$ & $78.7 \pm 1.4$ \\
Claude-4.5-Sonnet-thinking & $67.3 \pm 0.8$ & $62.1 \pm 1.1$ \\
Claude-4.5-Sonnet-no-thinking & $63.7 \pm 1.0$ & $58.1 \pm 0.9$ \\
Gemini-3-Pro-thinking & $65.0 \pm 1.2$ & $59.6 \pm 0.9$ \\
Gemini-3-Pro-no-thinking & $62.7 \pm 1.1$ & $55.7 \pm 1.3$ \\
GPT-5.2-thinking & $61.3 \pm 0.7$ & $54.6 \pm 1.3$ \\
GPT-5.2 & $58.5 \pm 1.5$ & $51.3 \pm 1.2$ \\
Doubao-seed-1-8-251215-thinking & $61.1 \pm 1.0$ & $56.6 \pm 0.8$ \\
Doubao-seed-1-8-251215 & $59.2 \pm 1.3$ & $53.6 \pm 1.1$ \\
DeepSeek-V3.2-thinking & $58.1 \pm 1.4$ & $46.5 \pm 0.8$ \\
DeepSeek-V3.2-no-thinking & $57.1 \pm 1.7$ & $45.4 \pm 1.5$ \\
Kimi-k2-thinking & $57.5 \pm 0.9$ & $54.4 \pm 1.1$ \\
Kimi-k2 & $54.6 \pm 1.2$ & $51.4 \pm 0.7$ \\
HunYuan-2.0-thinking & $57.3 \pm 1.1$ & $51.0 \pm 1.3$ \\
HunYuan-2.0-instruct & $52.7 \pm 1.6$ & $46.0 \pm 1.2$ \\
Grok-4.1-thinking & $52.7 \pm 0.8$ & $48.1 \pm 1.0$ \\
Grok-4.1 & $42.7 \pm 1.4$ & $42.7 \pm 1.8$ \\
Qwen-max-20250923-thinking & $52.5 \pm 0.9$ & $49.1 \pm 1.0$ \\
Qwen-max-20250923-no-thinking & $48.3 \pm 1.3$ & $45.2 \pm 0.9$ \\
GLM-4.7-thinking & $49.0 \pm 1.5$ & $42.5 \pm 1.6$ \\
GLM-4.7-no-thinking & $48.8 \pm 1.8$ & $40.9 \pm 1.4$ \\
\bottomrule
\end{tabular}
\end{table}

\begin{table*}[htbp]
\centering
\small
\caption{Data pool selection based on input state.}
\setlength{\tabcolsep}{6pt}
\begin{tabular}{@{}p{2.0cm}p{5.5cm}p{6.5cm}@{}}
\toprule
\textbf{Input State} & \textbf{Data Pool} & \textbf{Examples} \\
\midrule
Entity & Stocks and funds & Tencent, CATL, fund/ETF tickers \\
Time & Trading days, a period of time & 2025-11-19; FY2025\_Q3, 2024-10-05--2024-11-05 \\
Sector & Industry classifications, sector indices & New Energy Vehicles; Semiconductor; CSI 300 sector index \\
Event & Earnings releases, dividends, corporate announcements, ... & Q3 earnings release; dividend distribution; share buyback announcement \\
\bottomrule
\end{tabular}
\label{tab:data_pools_appendix}
\end{table*}

\section{State Transition Example}
\label{appendix:state_transition_example}
Table~\ref{tab:state_transition_examples} presents illustrative examples of valid state transitions within our framework. For each transition type, we demonstrate how the answer obtained at hop $k$ (highlighted in red) functions as a key constraint or subject for formulating the subsequent question at hop $k+1$.

\begin{table*}[htbp]
\centering
\small
\caption{State transition rules with examples. Each row shows a valid transition illustrated with consecutive hop questions. The ``Indicator'' state terminates the reasoning chain.}
\begin{tabular}{@{}p{1.8cm}p{2.2cm}p{5.5cm}p{5.5cm}@{}}
\toprule
\textbf{From State} & \textbf{To State} & \textbf{Hop $k$ Question} & \textbf{Hop $k$+1 Question} \\
\midrule
\multirow{4}{*}{Entity} 
& Entity 
& Question: Which fund holds the highest value of Moutai? \textit{(Answer Type: \textcolor{red}{Entity})} 
& Who is the fund manager of \textcolor{red}{that fund}? \textit{(Answer Type: \textcolor{red}{Entity})} \\
\cmidrule{2-4}
& Sector
& Among 4 companies, which had the highest net profit margin in Q3 2025? \textit{(Answer Type: \textcolor{red}{Entity})}
& What sector does \textcolor{red}{that company} belong to? \textit{(Answer Type: Sector)} \\
\cmidrule{2-4}
& Time 
& Among 4 companies, which had the highest net profit margin in Q3 2025? \textit{(Answer Type: \textcolor{red}{Entity})} 
& When was \textcolor{red}{that company} listed? \textit{(Answer Type: Time)} \\
\cmidrule{2-4}
& Indicator 
& What was CATL's closing price on Dec 30, 2025? \textit{(Answer Type: \textcolor{red}{Indicator})} 
& \textit{369.20 CNY (Terminal, Answer Type: Indicator)} \\
\midrule
\multirow{2}{*}{Time} 
& Entity 
& When did Jingjin Equipment Inc release its 2nd report in 2024? \textit{(Answer Type: \textcolor{red}{Time})} 
& \textcolor{red}{On that day}, which fund had weekly return $>$ -0.5\%? \textit{(Answer Type: Entity)} \\
\cmidrule{2-4}
& Indicator 
& Which day did Moutai's price peak in Nov 2025? \textit{(Answer Type: \textcolor{red}{Time})} 
& What was CATL's closing price on \textcolor{red}{that day}? \textit{(Answer Type: Indicator)} \\
\midrule
Sector & Entity 
& Which sector does CATL belong to? \textit{(Answer Type: \textcolor{red}{Sector})} 
& Which stock in \textcolor{red}{that sector} had highest 5-day gain? \textit{(Answer Type: Entity)} \\
\midrule
Event & Time 
& Kweichow Moutai dividend distribution \textit{(Answer Type: \textcolor{red}{Event})} 
& What was the date of \textcolor{red}{that announcement}? \textit{(Answer Type: Time)} \\
\midrule
Indicator & \textit{(Terminal)} & \multicolumn{2}{l}{\textit{Numerical values serve as final answers and cannot constrain subsequent queries.}} \\
\bottomrule
\end{tabular}
\label{tab:state_transition_examples}
\end{table*}

\section{Single-Hop Pattern Taxonomy}
\label{appendix:patterns}
We present a detailed taxonomy of 21 single-hop QA generation patterns in Table~\ref{tab:pattern_overview}, organized into three categories: Entity-Time Comparison, Event-Based, and Entity Relationship patterns. 

\begin{table*}[htbp]
    \centering
    \caption{Single-Hop QA Generation Patterns: Question Input $\rightarrow$ Answer Type.}
    \label{tab:pattern_overview}
    \small
    \setlength{\tabcolsep}{3pt}
    \renewcommand{\arraystretch}{1.4}
    \begin{tabular}{@{}clp{4.2cm}lp{5.8cm}@{}}
        \toprule
        \textbf{\#} & \textbf{Question} & \textbf{Condition} & \textbf{Answer} & \textbf{Example} \\
        \midrule
        \multicolumn{5}{l}{\textit{Entity-Time Comparison Patterns}} \\
        1 & ME + ST & Rank by indicator, select top-$n$ & SE & Which of Tencent, Xiaomi, Alibaba, CATL had highest volume on 2025-11-19? \\
        2 & ME + ST & Multi-indicator filtering (unique match) & SE & Which of Tencent, Xiaomi, Alibaba, CATL had volume $>$ 100M and turnover $<$ 1\% on 2025-11-19? \\
        3 & ME + ST & Compare quantitative indicator values & IND & What are net capital inflows for Tencent, Xiaomi, Alibaba on 2025-11-19? Which is largest? \\
        4 & SE + MT & Rank by indicator, select top-$n$ & ST & On which day did Tencent have highest volume among 2025-11-17/18/19? \\
        5 & SE + MT & Compare quantitative indicator values & IND & What was Tencent's volume on 2025-11-17/18/19? Which day was highest? \\
        6 & SE + MT & First time crossing special threshold & ST & When did Apple's stock first exceed \$150 among 2024-01/02/03? \\
        7 & SE + MT & Multi-indicator filtering (unique match) & ST & On which day among 2025-11-17/18/19 did Tencent have volume $>$ 100M and turnover $<$ 1\%? \\
        8 & SE + MT & Query financial report release date & ST & When was Tencent's report released among Q1/Q2/Q3 2023? \\
        9 & ME + ST & Rank by indicator, select top-$n$ & ME & Which two of Tencent, Xiaomi, Alibaba, CATL had highest volume on 2025-11-19? \\
        10 & SE + MT & Rank by indicator, select top-$n$ & MT & Which two days of 2025-11-17/18/19 had highest volume for Tencent? \\
        11 & ME + MT & Rank times by cross-entity percentile mean & MT & Among 2025-11-17/18/19, which two days had best average volume for Tencent, Xiaomi, Alibaba? \\
        12 & ME + MT & Rank entities by cross-time percentile mean & ME & Among Tencent, Xiaomi, Alibaba, which two had best average volume across 2025-11-17/18/19? \\
        13 & ME + MT & Filter by trend (growth/decline) & ME & Which of Tencent, Xiaomi, Alibaba showed continuous growth across 2025-11-17/18/19? \\
        \midrule
        \multicolumn{5}{l}{\textit{Event-Based Patterns}} \\
        14 & E + SE + NT & Extract time-related info & ST & When was Tencent founded? / When did BYD go public? \\
        15 & E + SE + NT & Query specific event date & ST & When was Tencent added to military list? \\
        16 & E + SE + NT & Locate by ordinal (1st/last/$n$-th) & ST & When was Tencent's first report release in 2024? \\
        17 & E + SE + NT & Locate multiple events (first $n$/all) & MT & What are all report release dates for Tencent in 2024? \\
        18 & E + NE + NT & Filter snippets, extract event date & ST & When did the Microsoft--Activision acquisition happen? \\
        \midrule
        \multicolumn{5}{l}{\textit{Entity Relationship Patterns}} \\
        19 & SS & Query stock's sector/industry & SEC & What sector does BYD belong to? \\
        20 & SEC & Query sector constituent stocks & MS & What stocks are in the NEV sector? \\
        21 & SEC & Query sector constituent funds/ETFs & MF & What ETFs are related to the NEV sector? \\
        \bottomrule
        \multicolumn{5}{l}{\footnotesize ME=Multi-Entity, SE=Single-Entity, MT=Multi-Time, ST=Single-Time, E=Event, NE=No-Entity,} \\
        \multicolumn{5}{l}{\footnotesize NT=No-Time, IND=Indicator, SS=Single-Stock, SEC=Sector, MS=Multi-Stock, MF=Multi-Fund} \\
    \end{tabular}
\end{table*}

\section{Guide For Chain-Merge Verification}
\label{appendix:Guide For Chain-Merge Verification}
We show the detailed guide for Chain-Merge Verification in Table~\ref{tab:chain_merge_verification}. For questions, criteria such as Intent Preservation and Logical Consistency maintain semantic integrity. They aim to eliminate ambiguity and prevent information leakage. For answers, criteria like Evidence Restriction and Faithfulness constrain outputs to the merged query. This ensures every intermediate answer is explicit and prohibits unsupported inferences. We provide concrete examples for each criterion to facilitate both automated and human evaluation.

\begin{table*}[t!] 
    \centering
    \caption{Guide For Chain-Merge Verification.}
    \label{tab:chain_merge_verification} 
    \small 
    \setlength{\tabcolsep}{4pt} 
    \begin{tabular}{@{}p{1.8cm}p{14cm}@{}}
        \hline 
        \multicolumn{2}{l}{\textbf{Question}} \\
        \hline
        Process Concealment & 
        \textbf{Description}: Intermediate results and final answers must not be revealed via direct mentions or contextual cues. \\
        & \textbf{Non-compliant Q}: "What was Apple's stock price on Dec 1, 2025, \textcolor{red}{given it reported higher net income} than NVIDIA in FY2026\_Q3?" \\
        & \textbf{Compliant Q}: "What was the stock price on Dec 1, 2025, of the company (Apple or NVIDIA) with higher net income in FY2026\_Q3?" \\
        \hline
        Intent Preservation & 
        \textbf{Description}: The semantic intent and information of both source questions must be preserved in the merged question. \\
        & \textbf{Source Q}: · "Which company had higher net income in FY26\_Q3: Apple or Nvidia?" · "What was Apple's stock price on Dec 1, 2025?" \\
        & \textbf{Non-compliant Q}: "Between Apple and Nvidia, what was the Dec 1, 2025 stock price of the company with \textcolor{red}{the better performance} in FY26\_Q3?" \\
        & \textbf{Compliant Q}: Between Apple and Nvidia, what was the Dec 1, 2025 stock price of the company with the higher net income in FY26\_Q3? \\
        \hline
        Disambiguation & 
        \textbf{Description}: Critical constituents in questions, including entities, times, and metrics, must be explicitly specified and unambiguous. \\
        & \textbf{Non-compliant Q}: "What was Apple's \textcolor{red}{income} last quarter?" \\
        & \textbf{Compliant Q}: "What was Apple's Net Income in FY26\_Q1?" \\
        \hline
        Closure & 
        \textbf{Description}: All entities and metrics must remain within the scope of the original questions, rejecting extensions. \\
        & \textbf{Source Q}: ·"Which company had higher net income in FY26\_Q3: Apple or Nvidia?" · "What was Apple's stock price on Dec 1, 2025?" \\
        & \textbf{Non-compliant Q}: "What was the stock price on Dec 1, 2025, for the company with higher FY26\_Q3 net income: Apple, Nvidia, or \textcolor{red}{Tesla}?" \\
        & \textbf{Compliant Q}: "What was the stock price on Dec 1, 2025, for the company with higher FY26\_Q3 net income: Apple or Nvidia?" \\
        \hline
        Logical Consistency & 
        \textbf{Description}: Causal links and premises must be reasonable and coherent. \\
        & \textbf{Non-compliant Q}: "How did the Federal Reserve's first rate hike \textcolor{red}{in 2025 drive the S\&P 500's rally in 2022}?" \\
        & \textbf{Compliant Q}: "How did Apple's stock perform following the Federal Reserve's first rate hike in 2025?" \\
        \hline
        \multicolumn{2}{l}{\textbf{Answer}} \\
        \hline
        Evidence Restriction & 
        \textbf{Description}: The final answer must derive strictly from the newly merged single-hop question. \\
        & \textbf{Source Q}: ·"Which company had the largest market cap in the S\&P 500 on Dec 29, 2025?" · "What was Nvidia's stock price change in November 2025?" \\
        & \textbf{Non-compliant Q}: "\textcolor{red}{Which company had the highest price in the S\&P 500 on December 29, 2025?}" \\
        & \textbf{Compliant Q}: "What was the stock price change in November 2025 of the company that had the largest market cap in the S\&P 500 on Dec 29, 2025?" \\
        \hline
        Completeness & 
        \textbf{Description}: All intermediate reasoning steps must be explicitly stated and supported by the step-wise answers. \\
        & \textbf{Merge Q}: "What was the stock price change in November 2025 of the company that had the largest market cap in the S\&P 500 on Dec 29, 2025?" \\
        & \textbf{Non-compliant A}: "\textcolor{red}{- \$25.49}" \\
        & \textbf{Compliant A}: "As of Dec 29, 2025, the company with the largest market cap in the S\&P 500 is Nvidia (NVDA). For November 2025, Nvidia’s stock price showed a significant decline: - \$25.49" \\
        \hline
        Faithfulness & 
        \textbf{Description}: Conclusions must be inferred solely from the provided evidence, avoiding hallucinated claims or unjustified refusals. \\
        & \textbf{Merge Q}: "Which company had the largest market cap in S\&P 500 on Dec 29, 2025, and what was its FY26\_Q3 revenue YoY growth?" \\
        & \textbf{Non-compliant A1}: "As of December 29, 2025, the company with the largest market cap in the S\&P 500 is Nvidia (NVDA). For its FY26\_Q3, Nvidia reported a revenue growth of 62\% year-over-year (YoY). \textcolor{red}{AI demand will sustain its steady growth in Q4.}" \\
        & \textbf{Non-compliant A2}: "\textcolor{red}{Insufficient information} to provide an answer." \\
        & \textbf{Compliant A}: "As of December 29, 2025, the company with the largest market cap in the S\&P 500 is Nvidia (NVDA). For its FY26\_Q3, Nvidia reported a revenue growth of 62\% year-over-year (YoY). " \\
        \hline
    \end{tabular}
\end{table*}

\subsection{Case Study}
\label{appendix:case study}
To investigate the underlying mechanisms of agentic reasoning, we conducted a qualitative analysis on the reasoning chains of all models (by thinking mode). Specifically, we sampled 20\% of successful instances ($s_{final}=1$) and 20\% of failed instances ($s_{final}=0$) from each difficulty level. This section summarizes the critical factors for success and categorizes the primary patterns of failure.

\subsubsection{Mechanisms of Success}
Our analysis reveals that solving multi-hop questions—particularly those exceeding 3 hops—requires a synergy of three core capabilities: strategic planning and decomposition, efficient tool utilization, robust context retention and self-correction. 

As illustrated in \hyperref[box:success-2hop]{Box 1} and \hyperref[box:success-5hop]{Box 2}, successful agents do not merely retrieve facts; they construct a logical roadmap. They maintain critical constraints across long horizons and autonomously trigger reflection mechanisms to correct deviations when intermediate steps yield implausible results.

\subsubsection{Taxonomy of Failures}
Despite possessing the capabilities mentioned above, agent performance degrades significantly as task complexity increases. By annotating and classifying the failed samples, we identified four primary failure modes within the reasoning chain.

\begin{enumerate}
    \item \textbf{Decomposition and Planning Failures (38\%):} This is the most prevalent structural error. Agents often exhibit logical sequencing errors, where the execution order of sub-goals is irrational, blocking subsequent steps. Furthermore, agents often lack the ability to refine their plans, leading to "dead ends" where they repetitively execute ineffective steps without strategic adjustment.
    
    \item \textbf{Tool Misuse (31\%):} This category involves the selection of inappropriate tools for the immediate sub-task. For instance, an agent might query a stock price tool instead of a corporate profile tool to find an IPO date. Additionally, failures occur due to improper usage, such as providing incorrect parameters or formats, which results in execution errors or null returns.
    
    \item \textbf{Hallucination Propagation (17\%):} When facing information gaps, agents may fabricate plausible but incorrect answers. This is particularly detrimental in multi-hop reasoning, as a single hallucinated intermediate step invalidates the entire subsequent chain.
    
    \item \textbf{Information Extraction Failures (14\%):} When tool outputs contain noise or irrelevant facts, models may misidentify these distractions as key clues. This misinterpretation causes the intermediate reasoning steps to deviate from the correct path.
\end{enumerate}

\clearpage
\onecolumn  
\definecolor{mygreen}{RGB}{95, 180, 130} 
\begin{tcolorbox}[
    title={Box 1: Success Case - 2 hop},
    colback=white,
    colframe=mygreen,
    coltitle=black,
    fonttitle=\bfseries,
    breakable,
    width=\linewidth,
    left=2mm, right=2mm, top=2mm, bottom=2mm,
    fontupper=\small\rmfamily,
    boxsep=1pt]

    \noindent  
    \begin{longtable}{@{} P{0.8cm} P{\dimexpr\linewidth-0.8cm-2\tabcolsep} @{}}
    \label{box:success-2hop}
        
        \textbf{System:} & \includegraphics[width=\linewidth]{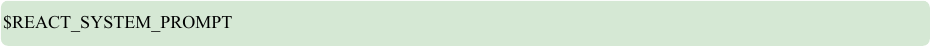}\\
        
        \textbf{User:} & \includegraphics[width=\linewidth]{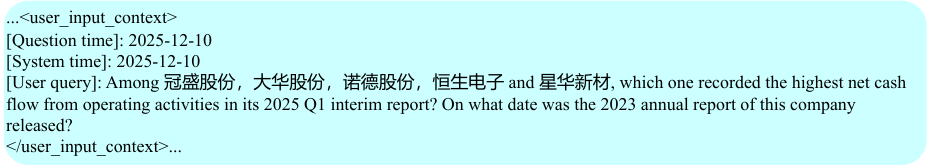}\\

        \textbf{Agent:} & \includegraphics[width=\linewidth]{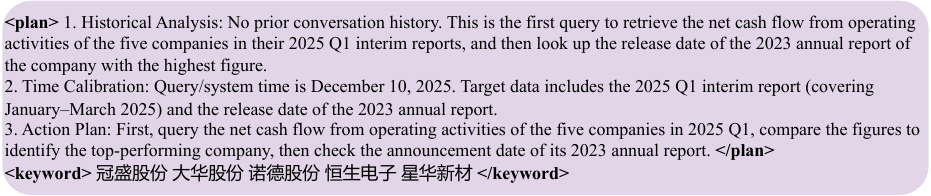} \\

        \textbf{User:} & \includegraphics[width=\linewidth]{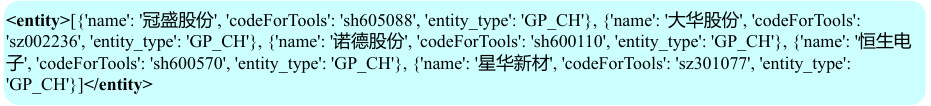} \\

        \textbf{Agent:} & \includegraphics[width=\linewidth]{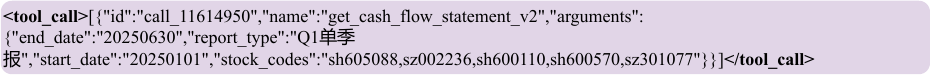} \\

        \textbf{Tool:} & \includegraphics[width=\linewidth]{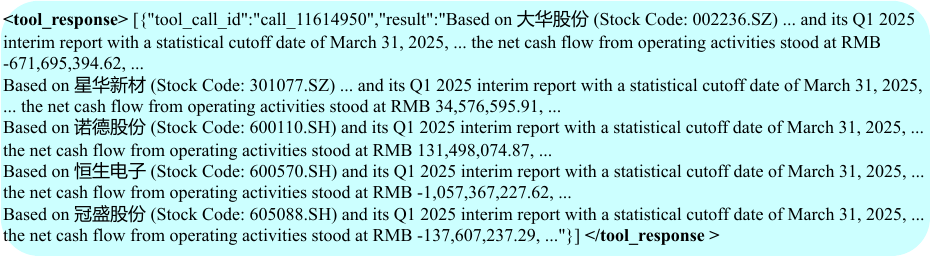} \\

        \textbf{Agent:} & \includegraphics[width=\linewidth]{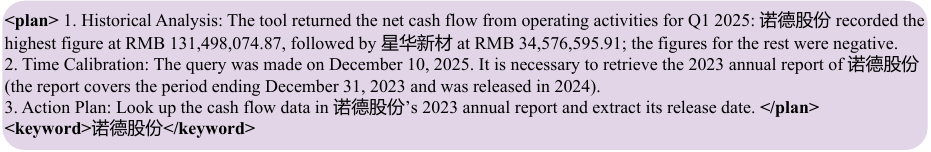} \\

        \textbf{User:} & \includegraphics[width=\linewidth]{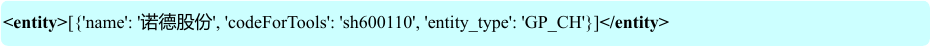} \\

        \textbf{Agent:} & \includegraphics[width=\linewidth]{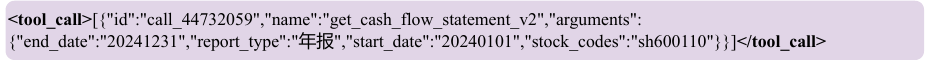} \\

        \textbf{Tool:} & \includegraphics[width=\linewidth]{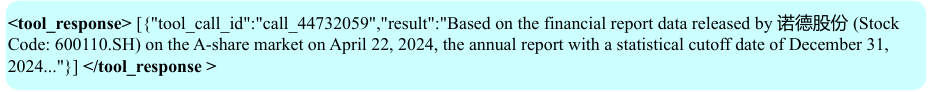} \\

        \textbf{Agent:} & \includegraphics[width=\linewidth]{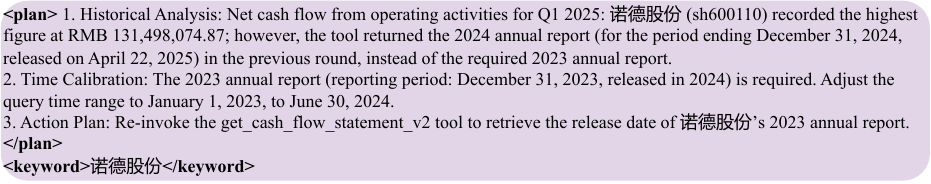} \\

        \textbf{User:} & \includegraphics[width=\linewidth]{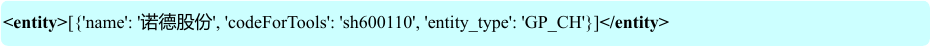} \\

        \textbf{Agent:} & \includegraphics[width=\linewidth]{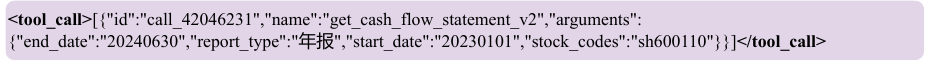} \\

        \textbf{Tool:} & \includegraphics[width=\linewidth]{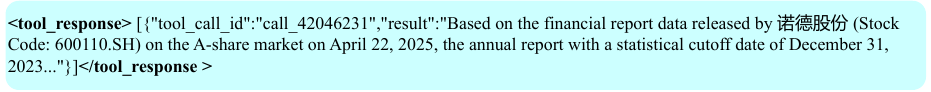} \\

        \textbf{Agent:} & \includegraphics[width=\linewidth]{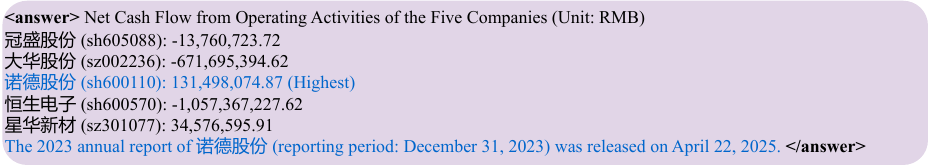} \\
        
    \end{longtable}

\end{tcolorbox}

\vspace{1\baselineskip}
\begin{tcolorbox}[
    title={Box 2: Success Case - 5 hop},
    colback=white,
    colframe=mygreen,
    coltitle=black,
    fonttitle=\bfseries,
    breakable,
    width=\linewidth,
    left=2mm, right=2mm, top=2mm, bottom=2mm,
    fontupper=\small\rmfamily,
    boxsep=1pt]

    \noindent  
    \begin{longtable}{@{} P{0.8cm} P{\dimexpr\linewidth-0.8cm-2\tabcolsep} @{}}
    \label{box:success-5hop}
        
        \textbf{System:} & \includegraphics[width=\linewidth]{figures/case1/1-1.pdf}\\
        
        \textbf{User:} & \includegraphics[width=\linewidth]{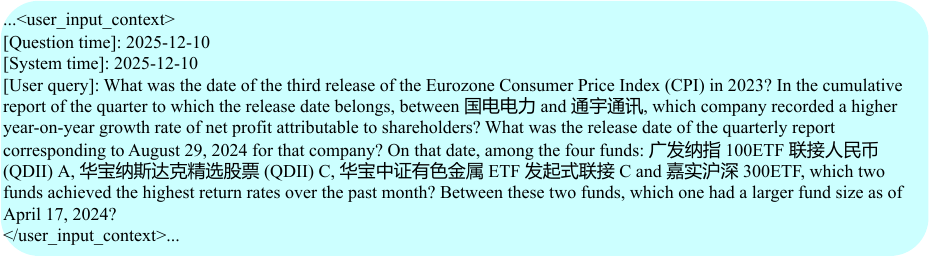}\\

        \textbf{Agent:} & \includegraphics[width=\linewidth]{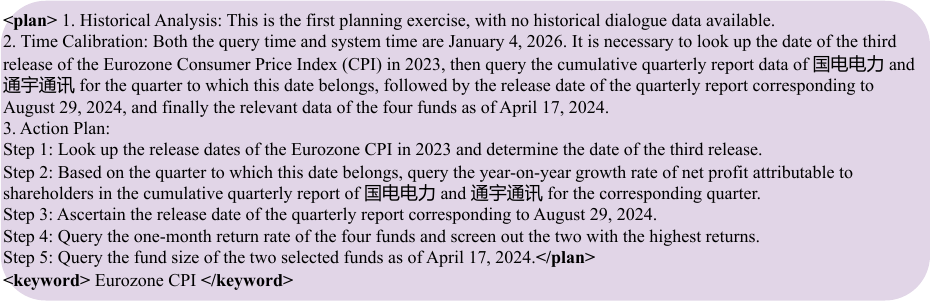} \\

        \textbf{User:} & \includegraphics[width=\linewidth]{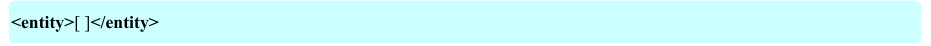} \\

        \textbf{Agent:} & \includegraphics[width=\linewidth]{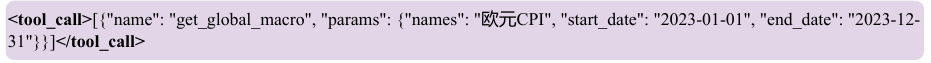} \\

        \textbf{Tool:} & \includegraphics[width=\linewidth]{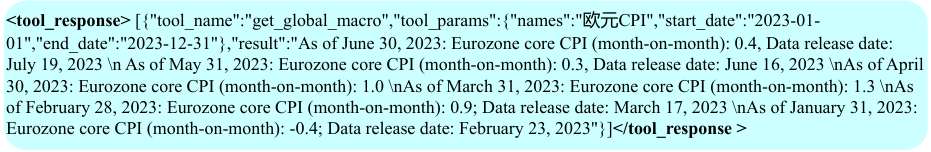} \\

        \textbf{Agent:} & \includegraphics[width=\linewidth]{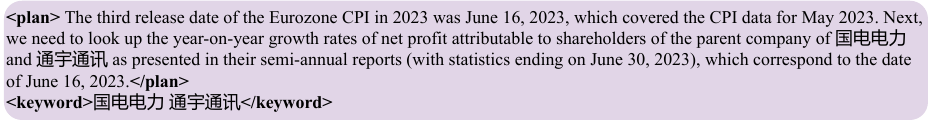} \\

        \textbf{User:} & \includegraphics[width=\linewidth]{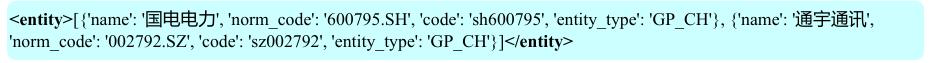} \\

        \textbf{Agent:} & \includegraphics[width=\linewidth]{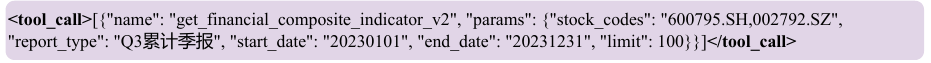} \\

        \textbf{Tool:} & \includegraphics[width=\linewidth]{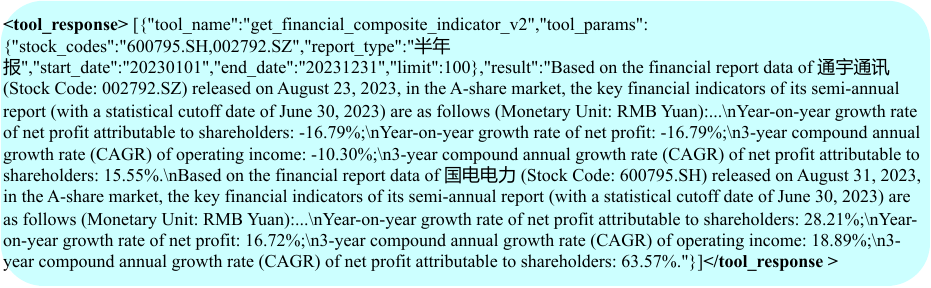} \\

        \textbf{Agent:} & \includegraphics[width=\linewidth]{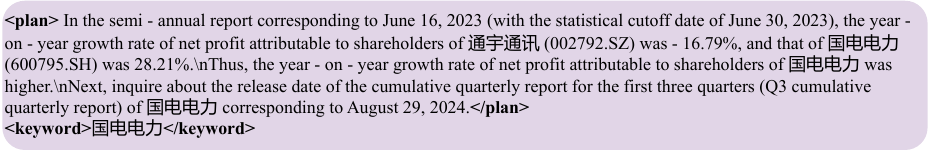} \\

        \textbf{User:} & \includegraphics[width=\linewidth]{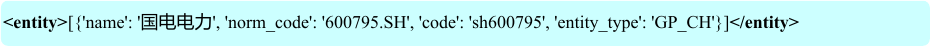} \\

        \textbf{Agent:} & \includegraphics[width=\linewidth]{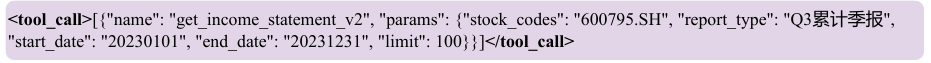} \\

        \textbf{Tool:} & \includegraphics[width=\linewidth]{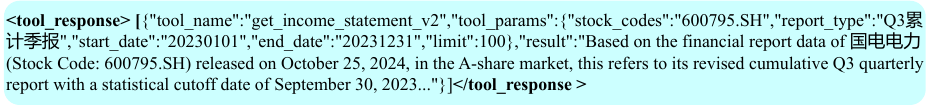} \\

        \textbf{Agent:} & \includegraphics[width=\linewidth]{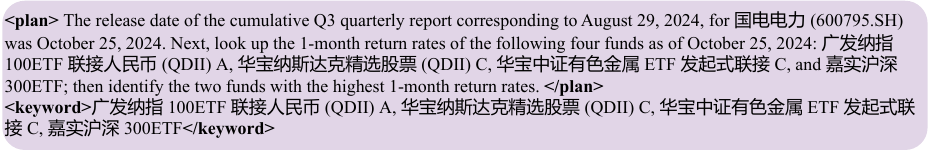} \\

        \textbf{User:} & \includegraphics[width=\linewidth]{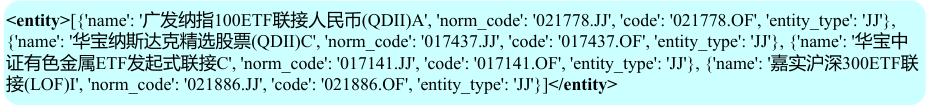} \\

        \textbf{Agent:} & \includegraphics[width=\linewidth]{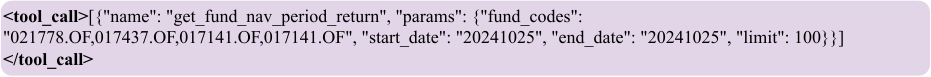} \\

        \textbf{Tool:} & \includegraphics[width=\linewidth]{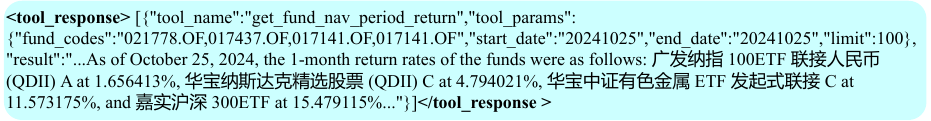} \\

        \textbf{Agent:} & \includegraphics[width=\linewidth]{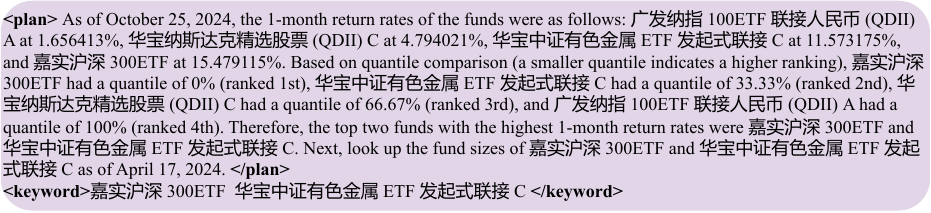} \\

        \textbf{User:} & \includegraphics[width=\linewidth]{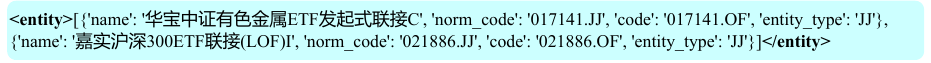} \\

        \textbf{Agent:} & \includegraphics[width=\linewidth]{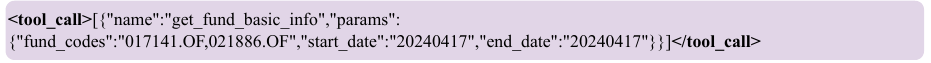} \\

        \textbf{Tool:} & \includegraphics[width=\linewidth]{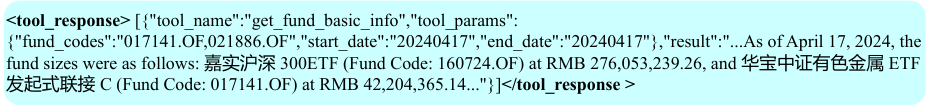} \\

        \textbf{Agent:} & \includegraphics[width=\linewidth]{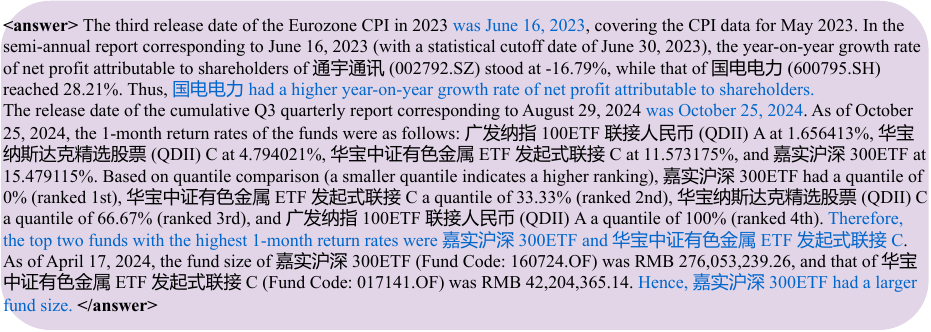} \\
    \end{longtable}

\end{tcolorbox}

\twocolumn  

\onecolumn
\section{Prompt of Single-Hop Pattern Taxonomy}
\label{appendix:prompts}

\begin{tcolorbox}[title={Pattern 1: ME+ST $\to$ SE (Rank by indicator, select top-$n$)},
    colback=blue!3,colframe=blue!50,coltitle=white,fonttitle=\bfseries\footnotesize,
    arc=2pt,boxrule=1pt,width=\textwidth,top=2pt,bottom=2pt]
\footnotesize
\begin{minipage}[t]{0.52\textwidth}
\textbf{[System Prompt]}\\[2pt]
\textcolor{red}{\textbf{Task Goal.}} Based on \textbf{entity list}, \textbf{time}, \textbf{tool names}, \textbf{tool descriptions}, and \textbf{tool retrieval results}, compare indicators across entities at the same time. Rank entities by a chosen indicator (higher/lower-is-better) and select the top-$n$. Question = \textbf{multi-entity + single-time}, answer = \textbf{single entity}.\\[2pt]
\textbf{Requirements:} Question must mention all entities. Answer derived from tool results only. Compare percentile levels.\\[2pt]
\textbf{Failure:} Output \texttt{\{\}} if 0 entities or invalid.
\end{minipage}%
\hfill
\begin{minipage}[t]{0.45\textwidth}
\textbf{Output:} \texttt{\{"question":"...", "answer":"...", "entities":[<single>]\}}\\[3pt]
\rule{\textwidth}{0.4pt}\\[3pt]
\textbf{[User Prompt]}\\
Entity List: \texttt{\{entities\}}\\
Time: \texttt{\{time\}}\\
Tool Name: \texttt{\{tool\_name\}}\\
Tool Description: \texttt{\{tool\_descriptions\}}\\
Tool Results: \texttt{\{tool\_result\}}
\end{minipage}
\end{tcolorbox}

\vspace{1.5mm}

\begin{tcolorbox}[title={Pattern 2: ME+ST $\to$ SE (Multi-indicator filtering, unique match)},
    colback=blue!3,colframe=blue!50,coltitle=white,fonttitle=\bfseries\footnotesize,
    arc=2pt,boxrule=1pt,width=\textwidth,top=2pt,bottom=2pt]
\footnotesize
\begin{minipage}[t]{0.52\textwidth}
\textbf{[System Prompt]}\\[2pt]
\textcolor{red}{\textbf{Task Goal.}} Based on inputs, select $n$ indicators. Find the \textbf{unique entity} satisfying a multi-condition filter (e.g., indicator A $>$ X AND indicator B $<$ Y). Question = \textbf{multi-entity + single-time}, answer = \textbf{unique entity} satisfying all conditions.\\[2pt]
\textbf{Requirements:} Question includes all entities and time. Exactly one entity satisfies conditions. Answer lists all indicator values.\\[2pt]
\textbf{Failure:} Output \texttt{\{\}} if no unique match.
\end{minipage}%
\hfill
\begin{minipage}[t]{0.45\textwidth}
\textbf{Output:} \texttt{\{"question":"...", "answer":"...", "entities":[<single>]\}}\\[3pt]
\rule{\textwidth}{0.4pt}\\[3pt]
\textbf{[User Prompt]}\\
Entity List: \texttt{\{entities\}}\\
Time: \texttt{\{time\}}\\
Tool Name: \texttt{\{tool\_name\}}\\
Tool Description: \texttt{\{tool\_descriptions\}}\\
Tool Results: \texttt{\{tool\_result\}}
\end{minipage}
\end{tcolorbox}

\vspace{1.5mm}

\begin{tcolorbox}[title={Pattern 3: ME+ST $\to$ IND (Compare quantitative indicator values)},
    colback=green!3,colframe=green!50,coltitle=white,fonttitle=\bfseries\footnotesize,
    arc=2pt,boxrule=1pt,width=\textwidth,top=2pt,bottom=2pt]
\footnotesize
\begin{minipage}[t]{0.52\textwidth}
\textbf{[System Prompt]}\\[2pt]
\textcolor{red}{\textbf{Task Goal.}} Select a \textbf{quantitative numeric indicator} (volume, return, turnover). Construct QA: question = \textbf{multi-entity + single-time}, answer focuses on \textbf{indicator values themselves} (max/min, top-N values), not just ``which entity''.\\[2pt]
\textbf{Requirements:} Question names all entities and time. Answer provides specific numeric values. \texttt{entities} field can be empty.\\[2pt]
\textbf{Failure:} Output \texttt{\{\}} if invalid comparison.
\end{minipage}%
\hfill
\begin{minipage}[t]{0.45\textwidth}
\textbf{Output:} \texttt{\{"question":"...", "answer":"...", "entities":[]\}}\\[3pt]
\rule{\textwidth}{0.4pt}\\[3pt]
\textbf{[User Prompt]}\\
Entity List: \texttt{\{entities\}}\\
Time: \texttt{\{time\}}\\
Tool Name: \texttt{\{tool\_name\}}\\
Tool Description: \texttt{\{tool\_descriptions\}}\\
Tool Results: \texttt{\{tool\_result\}}
\end{minipage}
\end{tcolorbox}

\vspace{1.5mm}

\begin{tcolorbox}[title={Pattern 4: SE+MT $\to$ ST (Rank by indicator, select top-$n$)},
    colback=orange!3,colframe=orange!50,coltitle=white,fonttitle=\bfseries\footnotesize,
    arc=2pt,boxrule=1pt,width=\textwidth,top=2pt,bottom=2pt]
\footnotesize
\begin{minipage}[t]{0.52\textwidth}
\textbf{[System Prompt]}\\[2pt]
\textcolor{red}{\textbf{Task Goal.}} Based on \textbf{entity}, \textbf{time list}, and \textbf{tool results}, compare a chosen indicator across different time points for the same entity. Rank times by indicator and select top-$n$. Question = \textbf{single-entity + multi-time}, answer = \textbf{single time}.\\[2pt]
\textbf{Requirements:} Question mentions entity and all times. Answer details indicator values at each time.\\[2pt]
\textbf{Failure:} Output \texttt{\{\}} if invalid question.
\end{minipage}%
\hfill
\begin{minipage}[t]{0.45\textwidth}
\textbf{Output:} \texttt{\{"question":"...", "answer":"...", "entities":[<time>]\}}\\[3pt]
\rule{\textwidth}{0.4pt}\\[3pt]
\textbf{[User Prompt]}\\
Entity List: \texttt{\{entities\}}\\
Time: \texttt{\{time\}}\\
Tool Name: \texttt{\{tool\_name\}}\\
Tool Description: \texttt{\{tool\_descriptions\}}\\
Tool Results: \texttt{\{tool\_result\}}
\end{minipage}
\end{tcolorbox}

\vspace{1.5mm}

\begin{tcolorbox}[title={Pattern 5: SE+MT $\to$ IND (Compare quantitative indicator values)},
    colback=orange!3,colframe=orange!50,coltitle=white,fonttitle=\bfseries\footnotesize,
    arc=2pt,boxrule=1pt,width=\textwidth,top=2pt,bottom=2pt]
\footnotesize
\begin{minipage}[t]{0.52\textwidth}
\textbf{[System Prompt]}\\[2pt]
\textcolor{red}{\textbf{Task Goal.}} Select a quantitative indicator. Ask for \textbf{indicator values at each time point} and provide comparison. Question = \textbf{single-entity + multi-time}, answer emphasizes \textbf{indicator values and comparison result}.\\[2pt]
\textbf{Requirements:} Question includes entity and all times. Answer gives specific numeric values for each time.\\[2pt]
\textbf{Failure:} Output \texttt{\{\}} if invalid question.
\end{minipage}%
\hfill
\begin{minipage}[t]{0.45\textwidth}
\textbf{Output:} \texttt{\{"question":"...", "answer":"...", "entities":[]\}}\\[3pt]
\rule{\textwidth}{0.4pt}\\[3pt]
\textbf{[User Prompt]}\\
Entity List: \texttt{\{entities\}}\\
Time: \texttt{\{time\}}\\
Tool Name: \texttt{\{tool\_name\}}\\
Tool Description: \texttt{\{tool\_descriptions\}}\\
Tool Results: \texttt{\{tool\_result\}}
\end{minipage}
\end{tcolorbox}

\newpage

\begin{tcolorbox}[title={Pattern 6: SE+MT $\to$ ST (First time crossing threshold)},
    colback=orange!3,colframe=orange!50,coltitle=white,fonttitle=\bfseries\footnotesize,
    arc=2pt,boxrule=1pt,width=\textwidth,top=2pt,bottom=2pt]
\footnotesize
\begin{minipage}[t]{0.52\textwidth}
\textbf{[System Prompt]}\\[2pt]
\textcolor{red}{\textbf{Task Goal.}} Choose a culturally meaningful threshold (e.g., 66, 88, 100, 888) and ask for the \textbf{first time} the indicator \textbf{crosses above/below} that threshold. Question = \textbf{single-entity + multi-time}, answer = \textbf{single time}.\\[2pt]
\textbf{Requirements:} Threshold should be meaningful. Ensure answer exists in data.\\[2pt]
\textbf{Failure:} Output \texttt{\{\}} if no crossing found.
\end{minipage}%
\hfill
\begin{minipage}[t]{0.45\textwidth}
\textbf{Output:} \texttt{\{"question":"...", "answer":"...", "entities":[<time>]\}}\\[3pt]
\rule{\textwidth}{0.4pt}\\[3pt]
\textbf{[User Prompt]}\\
Entity List: \texttt{\{entities\}}\\
Time: \texttt{\{time\}}\\
Tool Name: \texttt{\{tool\_name\}}\\
Tool Description: \texttt{\{tool\_descriptions\}}\\
Tool Results: \texttt{\{tool\_result\}}
\end{minipage}
\end{tcolorbox}

\vspace{1.5mm}

\begin{tcolorbox}[title={Pattern 7: SE+MT $\to$ ST (Multi-indicator filtering, unique match)},
    colback=orange!3,colframe=orange!50,coltitle=white,fonttitle=\bfseries\footnotesize,
    arc=2pt,boxrule=1pt,width=\textwidth,top=2pt,bottom=2pt]
\footnotesize
\begin{minipage}[t]{0.52\textwidth}
\textbf{[System Prompt]}\\[2pt]
\textcolor{red}{\textbf{Task Goal.}} Select $n$ suitable indicators from tool description and results. Find the \textbf{unique time} when indicator A $>$ threshold AND indicator B $<$ threshold. Question = \textbf{single-entity + multi-time}, answer = \textbf{unique time} satisfying all conditions.\\[2pt]
\textbf{Requirements:} Exactly one time satisfies conditions. Answer lists all indicator values at each time.\\[2pt]
\textbf{Failure:} Output \texttt{\{\}} if no unique match.
\end{minipage}%
\hfill
\begin{minipage}[t]{0.45\textwidth}
\textbf{Output:} \texttt{\{"question":"...", "answer":"...", "entities":[<time>]\}}\\[3pt]
\rule{\textwidth}{0.4pt}\\[3pt]
\textbf{[User Prompt]}\\
Entity List: \texttt{\{entities\}}\\
Time: \texttt{\{time\}}\\
Tool Name: \texttt{\{tool\_name\}}\\
Tool Description: \texttt{\{tool\_descriptions\}}\\
Tool Results: \texttt{\{tool\_result\}}
\end{minipage}
\end{tcolorbox}

\vspace{1.5mm}

\begin{tcolorbox}[title={Pattern 8: SE+MT $\to$ ST (Query financial report release date)},
    colback=orange!3,colframe=orange!50,coltitle=white,fonttitle=\bfseries\footnotesize,
    arc=2pt,boxrule=1pt,width=\textwidth,top=2pt,bottom=2pt]
\footnotesize
\begin{minipage}[t]{0.52\textwidth}
\textbf{[System Prompt]}\\[2pt]
\textcolor{red}{\textbf{Task Goal.}} From \textbf{tool retrieval results}, find the financial report release date. Construct a question asking for \textbf{the release date of a specific quarterly report}. Question = \textbf{single-entity + multi-time}, answer = \textbf{single time (release date)}.\\[2pt]
\textbf{Requirements:} Question must specify which quarterly report. Answer includes exact release date.\\[2pt]
\textbf{Failure:} Output \texttt{\{\}} if no release date found.
\end{minipage}%
\hfill
\begin{minipage}[t]{0.45\textwidth}
\textbf{Output:} \texttt{\{"question":"...", "answer":"...", "entities":[<time>]\}}\\[3pt]
\rule{\textwidth}{0.4pt}\\[3pt]
\textbf{[User Prompt]}\\
Entity List: \texttt{\{entities\}}\\
Time: \texttt{\{time\}}\\
Tool Name: \texttt{\{tool\_name\}}\\
Tool Description: \texttt{\{tool\_descriptions\}}\\
Tool Results: \texttt{\{tool\_result\}}
\end{minipage}
\end{tcolorbox}

\vspace{1.5mm}

\begin{tcolorbox}[title={Pattern 9: ME+ST $\to$ ME (Rank by indicator, select top-$n$)},
    colback=purple!3,colframe=purple!50,coltitle=white,fonttitle=\bfseries\footnotesize,
    arc=2pt,boxrule=1pt,width=\textwidth,top=2pt,bottom=2pt]
\footnotesize
\begin{minipage}[t]{0.52\textwidth}
\textbf{[System Prompt]}\\[2pt]
\textcolor{red}{\textbf{Task Goal.}} Compare indicators across entities at the same time. Rank entities by a chosen indicator and select the \textbf{top-$n$ entities} (where $n \geq 2$). Question = \textbf{multi-entity + single-time}, answer = \textbf{multiple entities}.\\[2pt]
\textbf{Requirements:} Question mentions all entities. Answer compares percentile levels and returns ordered list.\\[2pt]
\textbf{Failure:} Output \texttt{\{\}} if 0 entities or invalid.
\end{minipage}%
\hfill
\begin{minipage}[t]{0.45\textwidth}
\textbf{Output:} \texttt{\{"question":"...", "answer":"...", "entities":[<list>]\}}\\[3pt]
\rule{\textwidth}{0.4pt}\\[3pt]
\textbf{[User Prompt]}\\
Entity List: \texttt{\{entities\}}\\
Time: \texttt{\{time\}}\\
Tool Name: \texttt{\{tool\_name\}}\\
Tool Description: \texttt{\{tool\_descriptions\}}\\
Tool Results: \texttt{\{tool\_result\}}
\end{minipage}
\end{tcolorbox}

\vspace{1.5mm}

\begin{tcolorbox}[title={Pattern 10: SE+MT $\to$ MT (Rank by indicator, select top-$n$)},
    colback=teal!3,colframe=teal!50,coltitle=white,fonttitle=\bfseries\footnotesize,
    arc=2pt,boxrule=1pt,width=\textwidth,top=2pt,bottom=2pt]
\footnotesize
\begin{minipage}[t]{0.52\textwidth}
\textbf{[System Prompt]}\\[2pt]
\textcolor{red}{\textbf{Task Goal.}} Compare a chosen indicator across different time points for the same entity. Rank time points by indicator and select \textbf{top-$n$ times} (where $n \geq 2$). Question = \textbf{single-entity + multi-time}, answer = \textbf{multiple times}.\\[2pt]
\textbf{Requirements:} Question mentions entity and all times. Answer returns ordered time list.\\[2pt]
\textbf{Failure:} Output \texttt{\{\}} if invalid question.
\end{minipage}%
\hfill
\begin{minipage}[t]{0.45\textwidth}
\textbf{Output:} \texttt{\{"question":"...", "answer":"...", "entities":[<times>]\}}\\[3pt]
\rule{\textwidth}{0.4pt}\\[3pt]
\textbf{[User Prompt]}\\
Entity List: \texttt{\{entities\}}\\
Time: \texttt{\{time\}}\\
Tool Name: \texttt{\{tool\_name\}}\\
Tool Description: \texttt{\{tool\_descriptions\}}\\
Tool Results: \texttt{\{tool\_result\}}
\end{minipage}
\end{tcolorbox}

\newpage

\begin{tcolorbox}[title={Pattern 11: ME+MT $\to$ MT (Rank times by cross-entity percentile mean)},
    colback=red!3,colframe=red!50,coltitle=white,fonttitle=\bfseries\footnotesize,
    arc=2pt,boxrule=1pt,width=\textwidth,top=2pt,bottom=2pt]
\footnotesize
\begin{minipage}[t]{0.52\textwidth}
\textbf{[System Prompt]}\\[2pt]
\textcolor{red}{\textbf{Task Goal.}} For each time, compute the indicator's percentile mean (or equal-weighted average) across all entities as that day's composite score. Rank times and select top-$n$. Question = \textbf{multi-entity + multi-time}, answer = \textbf{multiple times}.\\[2pt]
\textbf{Requirements:} Question includes all entities and times. Answer details scores and returns ranked times.\\[2pt]
\textbf{Failure:} Output \texttt{\{\}} if invalid.
\end{minipage}%
\hfill
\begin{minipage}[t]{0.45\textwidth}
\textbf{Output:} \texttt{\{"question":"...", "answer":"...", "entities":[<times>]\}}\\[3pt]
\rule{\textwidth}{0.4pt}\\[3pt]
\textbf{[User Prompt]}\\
Entity List: \texttt{\{entities\}}\\
Time: \texttt{\{time\}}\\
Tool Name: \texttt{\{tool\_name\}}\\
Tool Description: \texttt{\{tool\_descriptions\}}\\
Tool Results: \texttt{\{tool\_result\}}
\end{minipage}
\end{tcolorbox}

\vspace{1.5mm}

\begin{tcolorbox}[title={Pattern 12: ME+MT $\to$ ME (Rank entities by cross-time percentile mean)},
    colback=red!3,colframe=red!50,coltitle=white,fonttitle=\bfseries\footnotesize,
    arc=2pt,boxrule=1pt,width=\textwidth,top=2pt,bottom=2pt]
\footnotesize
\begin{minipage}[t]{0.52\textwidth}
\textbf{[System Prompt]}\\[2pt]
\textcolor{red}{\textbf{Task Goal.}} For each entity, compute the indicator's percentile mean (or equal-weighted average) across all given times as that entity's composite score. Rank entities and select top-$n$. Question = \textbf{multi-entity + multi-time}, answer = \textbf{multiple entities}.\\[2pt]
\textbf{Requirements:} Question includes all entities and times. Answer details scores and returns ranked entities.\\[2pt]
\textbf{Failure:} Output \texttt{\{\}} if invalid.
\end{minipage}%
\hfill
\begin{minipage}[t]{0.45\textwidth}
\textbf{Output:} \texttt{\{"question":"...", "answer":"...", "entities":[<list>]\}}\\[3pt]
\rule{\textwidth}{0.4pt}\\[3pt]
\textbf{[User Prompt]}\\
Entity List: \texttt{\{entities\}}\\
Time: \texttt{\{time\}}\\
Tool Name: \texttt{\{tool\_name\}}\\
Tool Description: \texttt{\{tool\_descriptions\}}\\
Tool Results: \texttt{\{tool\_result\}}
\end{minipage}
\end{tcolorbox}

\vspace{1.5mm}

\begin{tcolorbox}[title={Pattern 13: ME+MT $\to$ ME (Filter by trend: growth/decline)},
    colback=red!3,colframe=red!50,coltitle=white,fonttitle=\bfseries\footnotesize,
    arc=2pt,boxrule=1pt,width=\textwidth,top=2pt,bottom=2pt]
\footnotesize
\begin{minipage}[t]{0.52\textwidth}
\textbf{[System Prompt]}\\[2pt]
\textcolor{red}{\textbf{Task Goal.}} Select an indicator and determine each entity's \textbf{trend} over the given time period (e.g., continuous growth, continuous decline, rise-then-fall, etc.). Find entities matching a specific trend. Question = \textbf{multi-entity + multi-time}, answer = \textbf{multiple entities} matching the trend.\\[2pt]
\textbf{Requirements:} Question asks ``which entities show [trend]?'' Answer lists indicator values across times.\\[2pt]
\textbf{Failure:} Output \texttt{\{\}} if no entities match.
\end{minipage}%
\hfill
\begin{minipage}[t]{0.45\textwidth}
\textbf{Output:} \texttt{\{"question":"...", "answer":"...", "entities":[<list>]\}}\\[3pt]
\rule{\textwidth}{0.4pt}\\[3pt]
\textbf{[User Prompt]}\\
Entity List: \texttt{\{entities\}}\\
Time: \texttt{\{time\}}\\
Tool Name: \texttt{\{tool\_name\}}\\
Tool Description: \texttt{\{tool\_descriptions\}}\\
Tool Results: \texttt{\{tool\_result\}}
\end{minipage}
\end{tcolorbox}

\vspace{1.5mm}

\begin{tcolorbox}[title={Pattern 14: E+SE+NT $\to$ ST (Extract time-related info)},
    colback=cyan!3,colframe=cyan!50,coltitle=white,fonttitle=\bfseries\footnotesize,
    arc=2pt,boxrule=1pt,width=\textwidth,top=2pt,bottom=2pt]
\footnotesize
\begin{minipage}[t]{0.52\textwidth}
\textbf{[System Prompt]}\\[2pt]
\textcolor{red}{\textbf{Task Goal.}} Based on \textbf{entity} and \textbf{tool results}, extract time-related information (founding date, listing date, etc.). Construct a QA pair: question = \textbf{single entity + no explicit time input}, answer = \textbf{single time}.\\[2pt]
\textbf{Requirements:} Answer must be exact date from tool results. No fabrication.\\[2pt]
\textbf{Failure:} Output \texttt{\{\}} if no time info available.
\end{minipage}%
\hfill
\begin{minipage}[t]{0.45\textwidth}
\textbf{Output:} \texttt{\{"question":"...", "answer":"...", "entities":[<time>]\}}\\[3pt]
\rule{\textwidth}{0.4pt}\\[3pt]
\textbf{[User Prompt]}\\
Entity List: \texttt{\{entities\}}\\
Tool Name: \texttt{\{tool\_name\}}\\
Tool Description: \texttt{\{tool\_descriptions\}}\\
Tool Results: \texttt{\{tool\_result\}}
\end{minipage}
\end{tcolorbox}

\vspace{1.5mm}

\begin{tcolorbox}[title={Pattern 15: E+SE+NT $\to$ ST (Query specific event date)},
    colback=cyan!3,colframe=cyan!50,coltitle=white,fonttitle=\bfseries\footnotesize,
    arc=2pt,boxrule=1pt,width=\textwidth,top=2pt,bottom=2pt]
\footnotesize
\begin{minipage}[t]{0.52\textwidth}
\textbf{[System Prompt]}\\[2pt]
\textcolor{red}{\textbf{Task Goal.}} From \textbf{tool results}, find dates of specific events (regulatory events, company milestones, business events, personnel changes, etc.). Construct a question asking for a \textbf{specific event date}. Question = \textbf{single entity}, answer = \textbf{single time}.\\[2pt]
\textbf{Requirements:} For periodic events, specify the year. Answer must be exact date.\\[2pt]
\textbf{Failure:} Output \texttt{\{\}} if no event date found.
\end{minipage}%
\hfill
\begin{minipage}[t]{0.45\textwidth}
\textbf{Output:} \texttt{\{"question":"...", "answer":"...", "entities":[<time>]\}}\\[3pt]
\rule{\textwidth}{0.4pt}\\[3pt]
\textbf{[User Prompt]}\\
Entity List: \texttt{\{entities\}}\\
Tool Name: \texttt{\{tool\_name\}}\\
Tool Description: \texttt{\{tool\_descriptions\}}\\
Tool Results: \texttt{\{tool\_result\}}
\end{minipage}
\end{tcolorbox}

\newpage

\begin{tcolorbox}[title={Pattern 16: E+SE+NT $\to$ ST (Locate by ordinal: 1st/last/$n$-th)},
    colback=cyan!3,colframe=cyan!50,coltitle=white,fonttitle=\bfseries\footnotesize,
    arc=2pt,boxrule=1pt,width=\textwidth,top=2pt,bottom=2pt]
\footnotesize
\begin{minipage}[t]{0.52\textwidth}
\textbf{[System Prompt]}\\[2pt]
\textcolor{red}{\textbf{Task Goal.}} From \textbf{tool results}, find periodic event dates (report releases, dividends, shareholder meetings, etc.). Construct a question using \textbf{ordinal keywords} (first, last, $n$-th) to locate a specific event date. Question = \textbf{single entity}, answer = \textbf{single time}.\\[2pt]
\textbf{Requirements:} Must use ordinal keywords. Need $\geq$2 same-type events for ranking.\\[2pt]
\textbf{Failure:} Output \texttt{\{\}} if $<$2 events found.
\end{minipage}%
\hfill
\begin{minipage}[t]{0.45\textwidth}
\textbf{Output:} \texttt{\{"question":"...", "answer":"...", "entities":[<time>]\}}\\[3pt]
\rule{\textwidth}{0.4pt}\\[3pt]
\textbf{[User Prompt]}\\
Entity List: \texttt{\{entities\}}\\
Tool Name: \texttt{\{tool\_name\}}\\
Tool Description: \texttt{\{tool\_descriptions\}}\\
Tool Results: \texttt{\{tool\_result\}}\\
Current Time: \texttt{\{current\_time\}}
\end{minipage}
\end{tcolorbox}

\vspace{1.5mm}

\begin{tcolorbox}[title={Pattern 17: E+SE+NT $\to$ MT (Locate multiple events: first $n$/all)},
    colback=cyan!3,colframe=cyan!50,coltitle=white,fonttitle=\bfseries\footnotesize,
    arc=2pt,boxrule=1pt,width=\textwidth,top=2pt,bottom=2pt]
\footnotesize
\begin{minipage}[t]{0.52\textwidth}
\textbf{[System Prompt]}\\[2pt]
\textcolor{red}{\textbf{Task Goal.}} From \textbf{tool results}, find multiple periodic event dates. Construct a question using keywords like ``first $n$'', ``last $n$'', or ``all'' to locate \textbf{multiple event dates}. Question = \textbf{single entity}, answer = \textbf{multiple times} ($\geq$2).\\[2pt]
\textbf{Requirements:} Answer must contain $\geq$2 times. For current year, add cutoff date constraint.\\[2pt]
\textbf{Failure:} Output \texttt{\{\}} if $<$2 events found.
\end{minipage}%
\hfill
\begin{minipage}[t]{0.45\textwidth}
\textbf{Output:} \texttt{\{"question":"...", "answer":"...", "entities":[<times>]\}}\\[3pt]
\rule{\textwidth}{0.4pt}\\[3pt]
\textbf{[User Prompt]}\\
Entity List: \texttt{\{entities\}}\\
Tool Name: \texttt{\{tool\_name\}}\\
Tool Description: \texttt{\{tool\_descriptions\}}\\
Tool Results: \texttt{\{tool\_result\}}\\
Current Time: \texttt{\{current\_time\}}
\end{minipage}
\end{tcolorbox}

\vspace{1.5mm}

\begin{tcolorbox}[title={Pattern 18: E+NE+NT $\to$ ST (Filter snippets, extract event date)},
    colback=cyan!3,colframe=cyan!50,coltitle=white,fonttitle=\bfseries\footnotesize,
    arc=2pt,boxrule=1pt,width=\textwidth,top=2pt,bottom=2pt]
\footnotesize
\begin{minipage}[t]{0.52\textwidth}
\textbf{[System Prompt]}\\[2pt]
\textcolor{red}{\textbf{Task Goal.}} Based on \textbf{event description} and \textbf{tool results}: (1) Filter relevant snippets matching the event description; (2) Extract the actual \textbf{event occurrence date} (not publish/report date); (3) Construct a question asking when the event occurred. Answer = \textbf{single time}.\\[2pt]
\textbf{Requirements:} Event date must be exact, not news publish time.\\[2pt]
\textbf{Failure:} Output \texttt{\{\}} if no actual event date found.
\end{minipage}%
\hfill
\begin{minipage}[t]{0.45\textwidth}
\textbf{Output:} \texttt{\{"question":"...", "answer":"...", "entities":[<time>]\}}\\[3pt]
\rule{\textwidth}{0.4pt}\\[3pt]
\textbf{[User Prompt]}\\
Event Description: \texttt{\{event\}}\\
Tool Name: \texttt{\{tool\_name\}}\\
Tool Description: \texttt{\{tool\_descriptions\}}\\
Tool Results: \texttt{\{tool\_result\}}
\end{minipage}
\end{tcolorbox}

\vspace{1.5mm}

\begin{tcolorbox}[title={Pattern 19: SS $\to$ SEC (Query stock's sector/industry)},
    colback=brown!3,colframe=brown!50,coltitle=white,fonttitle=\bfseries\footnotesize,
    arc=2pt,boxrule=1pt,width=\textwidth,top=2pt,bottom=2pt]
\footnotesize
\begin{minipage}[t]{0.52\textwidth}
\textbf{[System Prompt]}\\[2pt]
\textcolor{red}{\textbf{Task Goal.}} Based on \textbf{entity list} (single stock/company) and \textbf{tool results}, generate a single-hop QA pair: ``Stock $\to$ Sector/Industry''. Question asks which sector/industry the stock belongs to. Answer = \textbf{single sector/industry name}.\\[2pt]
\textbf{Requirements:} Answer must come from tool results. No fabrication.\\[2pt]
\textbf{Failure:} Output \texttt{\{\}} if sector info unavailable.
\end{minipage}%
\hfill
\begin{minipage}[t]{0.45\textwidth}
\textbf{Output:} \texttt{\{"question":"...", "answer":"...", "entities":[<sector>]\}}\\[3pt]
\rule{\textwidth}{0.4pt}\\[3pt]
\textbf{[User Prompt]}\\
Entity List: \texttt{\{entities\}}\\
Time: \texttt{\{time\}}\\
Tool Name: \texttt{\{tool\_name\}}\\
Tool Description: \texttt{\{tool\_descriptions\}}\\
Tool Results: \texttt{\{tool\_result\}}
\end{minipage}
\end{tcolorbox}

\vspace{1.5mm}

\begin{tcolorbox}[title={Pattern 20: SEC $\to$ MS (Query sector constituent stocks)},
    colback=brown!3,colframe=brown!50,coltitle=white,fonttitle=\bfseries\footnotesize,
    arc=2pt,boxrule=1pt,width=\textwidth,top=2pt,bottom=2pt]
\footnotesize
\begin{minipage}[t]{0.52\textwidth}
\textbf{[System Prompt]}\\[2pt]
\textcolor{red}{\textbf{Task Goal.}} Based on \textbf{entity list} (single sector/industry/concept) and \textbf{tool results}, generate a single-hop QA pair: ``Sector $\to$ Constituent Stocks''. Question asks which stocks belong to the sector. Answer = \textbf{multiple stock names}.\\[2pt]
\textbf{Requirements:} Answer must come from tool results. \texttt{entities} is a list of stock names.\\[2pt]
\textbf{Failure:} Output \texttt{\{\}} if constituent list unavailable.
\end{minipage}%
\hfill
\begin{minipage}[t]{0.45\textwidth}
\textbf{Output:} \texttt{\{"question":"...", "answer":"...", "entities":[<stocks>]\}}\\[3pt]
\rule{\textwidth}{0.4pt}\\[3pt]
\textbf{[User Prompt]}\\
Entity List: \texttt{\{entities\}}\\
Time: \texttt{\{time\}}\\
Tool Name: \texttt{\{tool\_name\}}\\
Tool Description: \texttt{\{tool\_descriptions\}}\\
Tool Results: \texttt{\{tool\_result\}}
\end{minipage}
\end{tcolorbox}

\newpage

\begin{tcolorbox}[title={Pattern 21: SEC $\to$ MF (Query sector constituent funds/ETFs)},
    colback=brown!3,colframe=brown!50,coltitle=white,fonttitle=\bfseries\footnotesize,
    arc=2pt,boxrule=1pt,width=\textwidth,top=2pt,bottom=2pt]
\footnotesize
\begin{minipage}[t]{0.52\textwidth}
\textbf{[System Prompt]}\\[2pt]
\textcolor{red}{\textbf{Task Goal.}} Based on \textbf{entity list} (single sector/industry/concept) and \textbf{tool results}, generate a single-hop QA pair: ``Sector $\to$ Constituent Funds/ETFs''. Question asks which funds/ETFs are related to the sector. Answer = \textbf{multiple fund/ETF names}.\\[2pt]
\textbf{Requirements:} Answer must come from tool results. \texttt{entities} is a list of fund/ETF names.\\[2pt]
\textbf{Failure:} Output \texttt{\{\}} if fund list unavailable.
\end{minipage}%
\hfill
\begin{minipage}[t]{0.45\textwidth}
\textbf{Output:} \texttt{\{"question":"...", "answer":"...", "entities":[<funds>]\}}\\[3pt]
\rule{\textwidth}{0.4pt}\\[3pt]
\textbf{[User Prompt]}\\
Entity List: \texttt{\{entities\}}\\
Time: \texttt{\{time\}}\\
Tool Name: \texttt{\{tool\_name\}}\\
Tool Description: \texttt{\{tool\_descriptions\}}\\
Tool Results: \texttt{\{tool\_result\}}
\end{minipage}
\end{tcolorbox}

\vspace{5mm}

\begin{tcolorbox}[colback=gray!5,colframe=gray!50,title={Pattern Color Legend},fonttitle=\bfseries\small]
\footnotesize
\begin{tabular}{@{}ll@{\hspace{15pt}}ll@{}}
\textcolor{blue!50}{\rule{10pt}{10pt}} ME+ST $\to$ SE & 
\textcolor{green!50}{\rule{10pt}{10pt}} ME+ST $\to$ IND &
\textcolor{orange!50}{\rule{10pt}{10pt}} SE+MT $\to$ ST/IND &
\textcolor{purple!50}{\rule{10pt}{10pt}} ME+ST $\to$ ME \\
\textcolor{teal!50}{\rule{10pt}{10pt}} SE+MT $\to$ MT &
\textcolor{red!50}{\rule{10pt}{10pt}} ME+MT $\to$ MT/ME &
\textcolor{cyan!50}{\rule{10pt}{10pt}} Event-based &
\textcolor{brown!50}{\rule{10pt}{10pt}} Entity Relationship \\
\end{tabular}

\vspace{3pt}
\textbf{Abbreviations:} ME=Multi-Entity, SE=Single-Entity, MT=Multi-Time, ST=Single-Time, E=Event, NE=No-Entity, NT=No-Time, IND=Indicator, SS=Single-Stock, SEC=Sector, MS=Multi-Stock, MF=Multi-Fund
\end{tcolorbox}
\vspace{1\baselineskip}

\section{Prompt of React Framework}
\label{appendix:Prompt of React Framework}
\begin{tcolorbox}[title={Prompt of React Framework},
    colback=mygreen3,colframe=mygreen2,coltitle=black,fonttitle=\bfseries\footnotesize,
    arc=2pt,boxrule=1pt,width=\textwidth,top=2pt,bottom=2pt,breakable]
\footnotesize
\noindent
\textbf{[System Prompt]}

\textbf{\# Role} \\
You are a high-IQ financial decision-making agent. Your task is to decide whether to invoke tools to obtain new information or summarize the final answer using existing information in the next step, based on the context, to solve the user's core problem.

\textbf{\# Special Instructions} \\
1. [Mandatory Memory Application]: ... \\
2. [Multi-round Execution]: ... \\
3. [Tool List]: ... \\
4. [Mandatory Tool Execution]: ...\\
   \quad - CRITICAL RULE: When the system provides you with information in the format of $\langle entity \rangle...\langle /entity \rangle$, it means you have obtained relevant entity information but no specific data yet. In this case, you must and only invoke tools (function call). \\
   \quad - Strictly prohibit outputting any analytical text. \\
   \quad - Strictly prohibit directly fabricating stock prices, financial report data, or answering questions solely based on entity codes. \\
   \quad - Example of Violation: After seeing "Jikai Co., Ltd.: sz002691", directly outputting "The current stock price of Jikai Co., Ltd. is..." (This is an absolutely forbidden hallucination).

\textbf{\# Output Format (Requirements)} \\
Please select the only corresponding output format below according to the current dialogue status:

\textbf{\#\# Status A: Planning Required (Start / Loop)} \\
If there is no $\langle entity \rangle$ information available currently and you need to query data: ...\\
(Stop output and wait for the system to return $\langle entity \rangle$ information)

\textbf{\#\# Status B: Entities Ready (Tool Execution)} \\
Trigger Condition: If and only if the input context contains the $\langle entity \rangle...\langle /entity \rangle$ information returned by the system. \\
In this status, you must invoke tools (function call) and are prohibited from outputting any other content.

\textbf{\#\# Status C: Sufficient Information (Answer)} \\
Trigger Condition: If and only if the data in historical dialogues completely covers all dimensions of the [User Input]. If the user's question involves multiple dimensions (e.g., multi-entity, multi-time comparison), you must obtain data for each dimension by invoking tools, without exception. \\
Important Notes \\
Unit Conversion Verification: In the results returned by tools, ...and strictly prohibit conclusion distortion caused by calculation or unit errors...\\
\noindent\rule{\linewidth}{0.4pt}
\vspace{4pt}
\textbf{[User Prompt]}

\textbf{\# Context Data}

User Input Context (Including Time Benchmark)

\textbf{<user\_input\_context>}

[Query Time]: \{query\_time\}

[System Time]: \{current\_time\}

[User Query]: \{user\_question\}

\textbf{</user\_input\_context>}

Notes:
- [Query Time]: Time entities (e.g., today, recently, etc.) are based on the query time.
- [System Time]: Current system time.

\textbf{\# Execution}
Now, based on the above information, output the next content in accordance with the **Output Format** specified in the System Prompt.

\end{tcolorbox}
\vspace{1\baselineskip}

\section{Prompt of Evaluation}
\label{appendix:Prompt of Evaluation}
\begin{tcolorbox}[title={Prompt of Final Answer Evaluation},
    colback=mygreen3,colframe=mygreen2,coltitle=black,fonttitle=\bfseries\footnotesize,
    arc=2pt,boxrule=1pt,width=\textwidth,top=2pt,bottom=2pt,breakable]
\footnotesize
\textbf{[System Prompt]}\\
\noindent\textbf{Evaluate the Student Answer objectively and rigorously against the provided Standard Answer.} You must abandon any tendency to solve the problem independently and strictly act as an execution tool for evaluating answers.

\vspace{4pt}
\noindent \textbf{Final Answer Scoring Rules (Evaluate Only the Final Answer, Ignore the Process)} \\
You are only allowed to check whether the Final Answer is consistent with the Standard Answer (ignore format differences such as units and percentage conversions). \\
- If consistent: Award 1 point. \\
- If inconsistent: Award 0 point. \\
No scores between 0 and 1 are permitted; no points shall be added or deducted for the process.

\vspace{4pt}
\noindent You must provide feedback in the following format strictly, with no redundant opening remarks: \\
1. [Scoring Basis]: State whether the final answers are consistent in no more than 150 words. \\
2. [JSON]: 
\begin{verbatim}
{"answer_score": score_value}
\end{verbatim}

\vspace{2pt}
\noindent The grading task officially begins:

\noindent\rule{\linewidth}{0.4pt}
\vspace{4pt}
\textbf{[User Prompt]}\\
<Question>:
\{prompt\}

<Standard Answer>:
\{response\_reference\}

<Student Answer>:
\{response\}

\end{tcolorbox}

\begin{tcolorbox}[title={Prompt of Process Evaluation},
    colback=mygreen3,colframe=mygreen2,coltitle=black,fonttitle=\bfseries\footnotesize,
    arc=2pt,boxrule=1pt,width=\textwidth,top=2pt,bottom=2pt,breakable]
\footnotesize
\textbf{[System Prompt]}\\
\noindent \textbf{Evaluate the Student Answer objectively and rigorously against the provided Standard Answer.} You must abandon any tendency to solve the problem independently and strictly act as an execution tool for evaluating answers.

\vspace{4pt}
\noindent Process Answer Scoring Rules (Evaluate Only the Process, Ignore the Final Answer Entirely) \\
The Standard Answer contains both "Process Answers" and "Final Answers" (or similar summaries/conclusions). You must: \\
- Extract and score only the process-related key points such as "process answers, intermediate steps, sub-question answers"; \\
- Explicitly exclude final conclusion content such as "final answers, final conclusions, therefore, in summary", which shall not be counted into N nor affect the score.

\vspace{4pt}
\noindent \textbf{Scoring Method (N Process Key Points, 0/1 Scoring for Each Item):} \\
1) Decompose the question and the corresponding key points of the Standard Answer, and count the total number of process key points in the Standard Answer as N (excluding the final answer). \\
2) For each process key point (which is the answer to an intermediate sub-question rather than supporting evidence), judge whether the Student Answer provides the corresponding and correct process conclusion/intermediate result: mark 1 for correct, 0 for incorrect or missing. \\
3) Full score for the process part is 0.5 points: Total Score = 0.5 * (Number of Hit Process Key Points) / N. If N=0, the total score shall be 0. \\
Evaluation Criterion: The Standard Answer is absolutely correct and indisputable. \\
If the Standard Answer requires "multiple choices/multiple key points", the student must cover all of them. Missing any sub-item will result in 0 score for that sub-item. \\
Compatibility: Substantial equality of numerical values shall be deemed correct (e.g., 0.15 = 15\%), and missing units shall not affect the correctness judgment.

\vspace{4pt}
\noindent You must provide feedback in the following format strictly, with no redundant opening remarks: \\
1. [Scoring Basis]: Briefly state the value of N, which process key points are hit and which are missed, within 150 words. \\
2. [JSON]: 
\begin{verbatim}
{"answer_score": score_value}
\end{verbatim}

\vspace{4pt}
\noindent\textbf{Example:} \\
\textbf{Question} \\
For the three companies Guqi Cashmere, Zhongda Leader, and Binhai Energy, in their Q3 single quarterly reports corresponding to October 16, 2024, which company had a net profit exceeding 10 million yuan and an income tax expense below 100,000 yuan? When was that company listed? On the listing day of that company, which fund between Harvest CSI Chuangke Chuangye 50 Index C and Tianhong Yongfeng Stable Pension Target 1-Year Holding Hybrid (FOF) had a 1-year return rate exceeding 10\% and a 3-month return rate being negative? \\
\textbf{Standard Answer} \\
In the Q3 single quarterly reports corresponding to October 16, 2024, Binhai Energy had a net profit of 1.8317 million yuan and an income tax expense of 1.3599 million yuan; Guqi Cashmere had a net profit of 40.6356 million yuan and an income tax expense of 0 yuan; Zhongda Leader had a net profit of 13.3423 million yuan and an income tax expense of 0.3693 million yuan. Only Guqi Cashmere met the conditions of net profit exceeding 10 million yuan (40.6356 million yuan > 10 million yuan) and income tax expense below 100,000 yuan (0 yuan < 100,000 yuan). Guqi Cashmere was listed on the A-share market on May 29, 2025. On May 29, 2025, Harvest CSI Chuangke Chuangye 50 Index C had a 1-year return rate of 12.66\% and a 3-month return rate of -9.01\%; Tianhong Yongfeng Stable Pension Target 1-Year Holding Hybrid (FOF) had a 1-year return rate of 2.69\% and a 3-month return rate of 0.86\%. Only Harvest CSI Chuangke Chuangye 50 Index C met the conditions of 1-year return rate exceeding 10\% and 3-month return rate being negative. \\
\textbf{Student Answer} \\
Only Guqi Cashmere met the conditions of net profit exceeding 10 million yuan (40.6356 million yuan > 10 million yuan) and income tax expense below 100,000 yuan (0 yuan < 100,000 yuan). Guqi Cashmere was listed on the A-share market on May 29, 2025. On May 29, 2025, only Harvest CSI Chuangke Chuangye 50 Index C met the conditions of 1-year return rate exceeding 10\% and 3-month return rate being negative.

\vspace{2pt}
\noindent [Scoring Basis] \\
N is 2. The student answer hit the two process key points of Guqi Cashmere and May 29, 2025, with no omissions. 0.5 * 2 / 2 = 0.5

[JSON]: 
\begin{verbatim}
{"answer_score": 0.5}
\end{verbatim}

\vspace{2pt}
\noindent The grading task officially begins:

\noindent\rule{\linewidth}{0.4pt}
\vspace{4pt}
\textbf{[User Prompt]}\\
<Question>:
\{prompt\}

<Standard Answer>:
\{response\_reference\}

<Student Answer>:
\{response\}

\end{tcolorbox}
\vspace{1\baselineskip}
\section{Prompt of Quality Control}
\label{appendix: Prompts of Quality Control}
\subsection{Single-Hop Verification}
\begin{tcolorbox}[title={Prompt of Single-Hop Verification},
    colback=mygreen3,colframe=mygreen2,coltitle=black,fonttitle=\bfseries\footnotesize,
    arc=2pt,boxrule=1pt,width=\textwidth,top=2pt,bottom=2pt,breakable]
\footnotesize
\textbf{[System Prompt]}\\
You are a professional answer comparison specialist, adept at determining whether the original answer can be directly derived from the reference materials.

\noindent\rule{\linewidth}{0.4pt}
\vspace{4pt}
\textbf{[User Prompt]}\\
\noindent Please strictly determine whether the original answer can be \textbf{directly derived} from the reference materials. Key information such as numerical values, percentages, dates and units \textbf{must be exactly consistent}.

\vspace{4pt}
\noindent Question: \{query\} \\
Original Answer: \{original\_answer\} \\
Reference Materials: \{content\}

\vspace{4pt}
\noindent Please return \textbf{only a JSON object} in the following format: 
\begin{verbatim}
{
  "can_derive": true/false,
  "reason": "Briefly state the reason"
}
\end{verbatim}

\vspace{4pt}
\noindent Notes: \\
1. If the core information of the original answer can be directly obtained from the reference materials, return `true` even if the expression is different. \\
2. Return `false` if there are discrepancies in numerical values, dates or key information. \\
3. Return `false` if the question does not contain explicit time information. \\
4. Do not include any explanatory text; return only the JSON object.
\end{tcolorbox}

\subsection{Chain-Merge Verification}

\begin{tcolorbox}[title={Prompt of Question Critic},
    colback=mygreen3,colframe=mygreen2,coltitle=black,fonttitle=\bfseries\footnotesize,
    arc=2pt,boxrule=1pt,width=\textwidth,top=2pt,bottom=2pt,breakable]

\footnotesize
\textbf{[System Prompt]}\\
\noindent \textbf{\# Task Objective}

\vspace{2pt}
\noindent Validate and, if necessary, optimize and rewrite the "draft merged question" to ensure that:\\
    - The defined scope of entities and time periods is strictly retained and explicitly presented, without expanding to the entire market or ambiguous sets;\\
    - No information in the original question is lost (including entities, time periods, indicators, comparison relationships, etc.);\\
    - No answers are revealed in advance, and only the question is posed;\\
    - The language is natural, uniquely solvable, and answerable using the provided data.\\

\vspace{2pt}
\noindent \textbf{\# Usage Instructions}

\vspace{2pt}
\noindent You will receive the following inputs:\\
    - Original QA pair (\texttt{original\_qa})\\
    - QA pair to be merged (\texttt{new\_qa})\\
    - Draft question text appended after the phrase "Draft Merged Question:" at the bottom of the message

\vspace{2pt}
\noindent \textbf{\# Notes}

\vspace{2pt}
\noindent \textbf{\#\# Core Principles}\\
    - The preceding question must not reveal the answer to the subsequent question, and vice versa.\\
    - ...

\vspace{2pt}
\noindent \textbf{\#\# Reference Specification}\\
    - Prohibit the use of references that cause scope expansion, e.g., "the company with the highest trading volume on 2025-11-10".\\
    - ...

\vspace{2pt}
\noindent \textbf{\#\# Information Completeness}\\
    - Do not add, delete, or weaken any existing entities.\\
    - ...

\vspace{2pt}
\noindent \textbf{\#\# Uniqueness and Solvability}\\
    - The question must be unique, solvable, and have a controllable search scope.\\
    - ...

\vspace{2pt}
\noindent \textbf{\#\# Expression Specification}\\
    - Avoid redundant sub-questions (content already described in the previous question does not need to be repeated).\\
    - ...

\vspace{2pt}
\noindent \textbf{\# Recommended Multi-step Splicing Structure}

\vspace{2pt}
    - In the first step, pose the question using an explicit set of entities.\\
    - In the second step, use "Among the entity/time set from the previous question, the company/fund/market/day/quarter/year with... " to connect, and make comparisons with newly added entities or conditions.\\
    - In subsequent steps, use "this company/that fund/that market/that day/that quarter/that year..." to build on the results of the previous step for further comparisons.\\
    - Maintain a non-expanding scope, unambiguous expression, and no answer leakage throughout the process.

\vspace{2pt}
\noindent \textbf{\# Correct Example}
\begin{quote}
    Which company had the highest trading volume on 2025-11-10, Giant Network or Century Huatong? \\
    Between the company with higher trading volume among Giant Network and Century Huatong and YooZoo Network, which one had a larger stock price increase on November 12, 2025? \\
    Between this company and Industrial and Commercial Bank of China, Daqin Railway, and Tongyi Co., Ltd., which one had the highest turnover rate on 2025-11-18?
\end{quote}

\vspace{2pt}
\noindent \textbf{\# Counterexample Explanations}

\vspace{2pt}
\noindent \textbf{Counterexample 1: Answer Leakage (Directly revealing intermediate answers)} \\
\textcolor{red}{Incorrect:} "...the company with higher transaction volume... (answer: Century Huatong) among [Century Huatong, Industrial and Commercial Bank of China...]..."\\
\textcolor{green}{Correction:} Use referential terms such as "that company" or "this company" for connection, without disclosing the specific entity name.

\vspace{5pt}
\noindent \textbf{Counterexample 2: Time Answer Leakage} \\
\textcolor{red}{Incorrect:} "Among the two days of 2025-05-22 and 2025-03-18 for Bubugao, on the day with higher trading volume... on May 22, 2025..."\\
\textcolor{green}{Correction:} Specific dates should not be mentioned in subsequent parts; use "the financial report quarter corresponding to that day" instead.

\vspace{5pt}
\noindent \textbf{Counterexample 3: ...}

\vspace{4pt}
\noindent \textbf{\# Output Format}

\vspace{2pt}
\noindent Return only the JSON object:
\begin{verbatim}
{
  "question": "<Optimized question or original question>"
}
\end{verbatim}
\noindent\rule{\linewidth}{0.4pt}
\vspace{4pt}
\textbf{[User Prompt]}\\
Original QA Pair:\\
\{original\_qa\}\\
QA Pair to Be Merged:\\
\{new\_qa\}

\end{tcolorbox}
\vspace{1\baselineskip}

\begin{tcolorbox}[title={Prompt of Answer Critic},
    colback=mygreen3,colframe=mygreen2,coltitle=black,fonttitle=\bfseries\footnotesize,
    arc=2pt,boxrule=1pt,width=\textwidth,top=2pt,bottom=2pt,breakable]

\footnotesize
\textbf{[System Prompt]}\\
\noindent \textbf{\# Task Objective}

\vspace{2pt}
\noindent Validate the \textit{draft merged answer} to ensure that: \\
- All factual details from the original QA pair and the QA pair to be merged are fully retained without omission or tampering; \\
- The answer responds smoothly to the merged question with coherent logic; \\
- The \texttt{entities} field correctly points to the answer entities/time periods corresponding to the last sub-question.

\vspace{2pt}
\noindent \textbf{\# Usage Instructions}

\vspace{2pt}
\noindent You will receive the following inputs:\\
    - Original QA pair (\texttt{original\_qa})\\
    - QA pair to be merged (\texttt{new\_qa})\\
    - Draft question text appended after the phrase "Draft Merged Question:" at the bottom of the message

\vspace{4pt}
\noindent \textbf{\# Validation Rules}

\vspace{2pt}
\noindent \textbf{\#\# Completeness Validation} \\
- All details (entities, numerical values, time periods, rankings, etc.) in the answer of the original QA pair must be fully retained; \\
- All details in the answer of the QA pair to be merged must be fully retained; \\
- No information from any sub-answer is allowed to be omitted.

\vspace{2pt}
\noindent \textbf{\#\# Accuracy Validation} \\
- All content must be directly derived from the original QA pair or the QA pair to be merged; fabrication or speculation is strictly prohibited; \\
- Numerical values, entity names, time periods, etc., must be consistent with the original data; \\
- Tampering with or misrepresenting the meaning of the original answers is prohibited.

\vspace{4pt}
\noindent \textbf{\# Output Format}

\vspace{2pt}
\noindent Return only the JSON object:
\begin{verbatim}
{
  "answer": "<Optimized answer or original answer>"
}
\end{verbatim}
\noindent\rule{\linewidth}{0.4pt}
\vspace{4pt}
\textbf{[User Prompt]}\\
Original QA Pair:\\
\{original\_qa\}\\
QA Pair to Be Merged:\\
\{new\_qa\}

\end{tcolorbox}

\end{document}